\documentclass[acmsmall, screen, nonacm]{acmart}

\AtBeginDocument{%
  }

\usepackage{graphicx}
\usepackage{epigraph}
\usepackage{tikz}
\usepackage[edges]{forest}
\usepackage{caption}
\usepackage{subcaption}

\begin{document}

\title{Creativity in AI: Progresses and Challenges}

\author{Mete Ismayilzada*}
\email{mahammad.ismayilzada@epfl.ch}
\affiliation{%
  \institution{EPFL, Università della Svizzera italiana (USI)}
  \country{Switzerland}
}

\author{Debjit Paul}
\email{debjit.paul@epfl.ch}
\affiliation{%
  \institution{EPFL}
  \country{Switzerland}
}

\author{Antoine Bosselut}
\email{antoine.bosselut@epfl.ch}
\affiliation{%
  \institution{EPFL}
  \country{Switzerland}
}

\author{Lonneke van der Plas*}
\email{lonneke.vanderplas@usi.ch}
\affiliation{%
  \institution{Università della Svizzera italiana (USI)}
  \country{Switzerland}
}

\begingroup
\renewcommand\thefootnote{\textsuperscript{*}}\footnotetext{Work done while at Idiap Research Institute}
\endgroup

\renewcommand{\shortauthors}{Ismayilzada et al.}

\begin{abstract}
Creativity is the ability to produce novel, useful, and surprising ideas, and has been widely studied as a crucial aspect of human cognition. Machine creativity on the other hand has been a long-standing challenge. With the rise of advanced generative artificial intelligence (AI), there has been renewed interest and debate regarding AI's creative capabilities. Therefore, it is imperative to revisit the state of creativity in AI and identify key progresses and remaining challenges.
In this work, we survey leading works studying the creative capabilities of AI systems, focusing on creative problem-solving, linguistic, artistic, and scientific creativity. 
Our review suggests that while the latest AI models are largely capable of producing linguistically and artistically creative outputs such as poems, images, and musical pieces, they struggle with tasks that require creative problem-solving, abstract thinking and compositionality and their generations suffer from a lack of diversity, originality, long-range incoherence and hallucinations. We also discuss key questions concerning copyright and authorship issues with generative models.  Furthermore, we highlight the need for a comprehensive evaluation of creativity that is process-driven and considers several dimensions of creativity. Finally, we propose future research directions to improve the creativity of AI outputs, drawing inspiration from cognitive science and psychology.

\end{abstract}

\maketitle

\section{Introduction}
\label{sec:introduction}

\epigraph{Computers can’t create anything. For creation requires, minimally, originating something. But computers originate nothing; they merely do that which we order them, via programs, to do.}{\textit{Ada Lovelace}}

Creativity, the ability to produce novel, useful, and surprising ideas \cite{boden2004creative}, is one of the major hallmarks of human intelligence \cite{guilford1967nature}. Since the invention of the first known general-purpose mechanical computer (known as Analytical Engine) designed by Babbage \cite{babbage1864}, the question of whether machines can truly think or create anything new has intrigued the scientific community \cite{wang2024aicreativehumans, turing1950, Newell1959ThePO}. Ada Lovelace, recognized as the first programmer by many, famously stated that the Analytical Engine \textit{has no pretensions to originate anything} \cite{Menabrea2015SketchOT} and Alan Turing, who laid the foundations of computer science, asserted that \textit{machines can never take us by surprise} \cite{turing1950}. Nevertheless, alongside the development of personal computers and advancements in Artificial Intelligence (AI), several symbolic-based and stochastic approaches were developed to endow machines with story generation \cite{Meehan1977TALESPINAI, Lebowitz1983CreatingAS, Prez2001MEXICAAC, Turner1994TheCP}, poetry writing \cite{masterman1971, Racter1984ThePB} and music composition skills \cite{brooks1957experiment, hiller1953musical}. However, these early approaches could not generalize beyond a set of limited domains \cite{Ji2020ACS, yao2019plan, Colton2012FullFACEPG}. 

Fast-forward to now, the advent of the Transformer architecture \cite{Vaswani2017AttentionIA} and the development of large language models (LLMs) \cite{zhao2023survey} in the past decade ushered a new age of intelligent systems with remarkable generative, reasoning, coding, mathematical and multimodal capabilities \cite{geminiteam2024gemini, bubeck2023sparks, wei2022emergent}. Transformer-based models can now produce long stories in various domains \cite{yang-etal-2022-re3, yao2019plan}, write poems about diverse topics \cite{Chakrabarty2022HelpMW, Ormazabal2022PoeLMAM}, compose high-fidelity songs \cite{Dhariwal2020JukeboxAG}, generate impressive high-quality images and videos \cite{videoworldsimulators2024, BetkerImprovingIG} and discover new scientific knowledge \cite{Jumper2021}.

\begin{figure}[t]
    \centering   
    \includegraphics[scale=1.0,width=\linewidth]{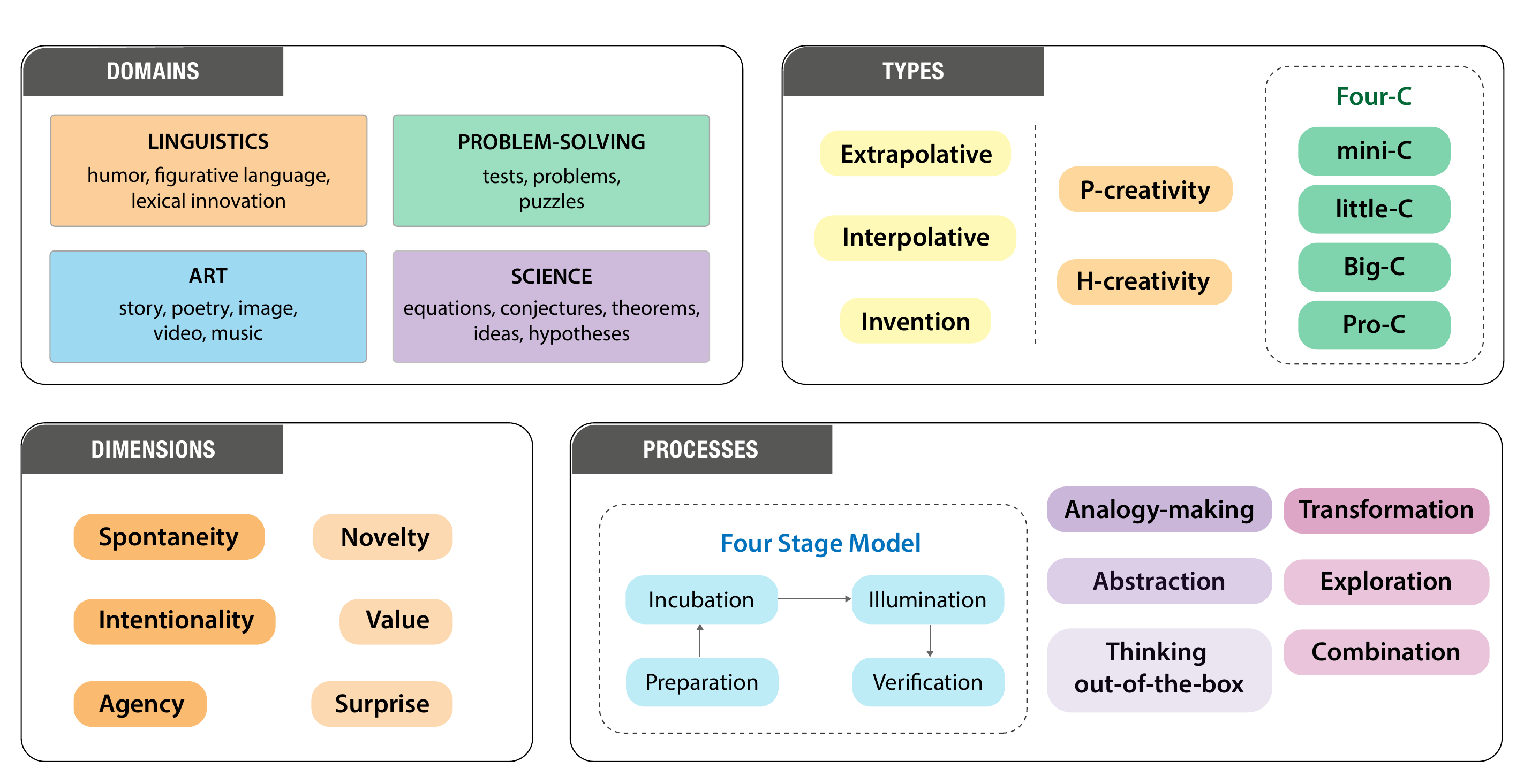}
    \caption{A summary of domains, dimensions, types and processes of creativity covered in this survey.}
    \label{fig:main-figure}
\end{figure}

While these remarkable achievements can be seen as signs of the presence of creative capacity in transformer-based language models, it should be noted that these models rely on an astronomically large number of parameters and are trained on massive amounts of public and private data \cite{Brown2020LanguageMA}. Hence, it is not entirely clear whether the seemingly extraordinary outputs of these models are the result of a truly creative inner process and robust generalization or the result of powerful interpolation and strong memorization skills \cite{McCoy2023EmbersOA, Hupkes2022ATA, Carlini2022QuantifyingMA, Bender2021OnTD, Marcus2020TheND, Bender2020ClimbingTN}. Recent works have shown that language models fail at real-world commonsense reasoning and compositionality tasks \cite{Ismayilzada2023CRoWBC, Dziri2023FaithAF}, occasionally copy large amounts of text from their training data \cite{lu2024aihumanityssalieriquantifying, McCoy2021HowMD}, significantly lag behind humans in creative writing \cite{ismayilzada2024evaluatingcreativeshortstory, Chakrabarty2023ArtOA}, produce less diverse content \cite{anderson2024, Padmakumar2023DoesWW}, struggle with creative problem-solving \cite{huang-etal-2024-lateval, Tian2023MacGyverAL} and abstract reasoning \cite{Gendron2023LargeLM, Mitchell2023ComparingHG} and suffer from factual inconsistency and hallucination issues \cite{banerjee2024llmshallucinateneedlive, Elazar2021MeasuringAI}. While previous works have reviewed developments on some aspects of AI creativity \cite{amin-burghardt-2020-survey, Lai2024ASO, Oliveira2017ASO, Elzohbi2023CreativeDG, Franceschelli2021CreativityAM, Rowe1993CreativityAS}, a more holistic and broader review of the field is necessary to understand its rapid advancements.


In this work, we provide the general AI audience with a timely summary of the state of the creative capabilities of the latest AI systems.  While creativity is a broad concept that can be explored in a wide range of areas, in this survey, we focus on four main areas where machine creativity has been extensively investigated: \textbf{Linguistic Creativity} (\S \ref{sec:linguistic_creativity}), \textbf{Creative Problem-Solving} (\S \ref{sec:problem_solving}), \textbf{Artistic Creativity} (\S \ref{sec:artistic-creativity}) and \textbf{Scientific Creativity} (\S \ref{sec:scientific-creativity}). These areas capture four crucial pillars of creative thinking in humans: linguistic creativity enables us to manipulate language in novel ways for effective communication of ideas; creative problem-solving helps us find efficient solutions by thinking out-of-the-box; artistic creativity allows for the expression of emotions, ideas, and aesthetics through various media; and scientific creativity drives innovation by enabling the formulation of new hypotheses, theories, and discoveries. Together, these areas represent distinct yet interconnected facets of creativity, providing a comprehensive framework for studying how machines can emulate or assist human-like creative processes across different domains. For each area, we survey representative tasks, resources, methods, and major findings and present a taxonomy of these works in Figure \ref{fig:creativity_taxonomy}. Our review indicates that although the latest AI models are generally proficient in generating linguistically and artistically creative outputs, such as poems, images, and music, they face challenges with tasks demanding creative problem-solving, abstract reasoning, and compositionality. Their outputs often lack diversity and originality, exhibit long-term incoherence, and are prone to hallucinations. We also briefly discuss the emerging challenges brought by generative models concerning \textbf{copyright} and \textbf{authorship} of artworks (\S \ref{sec:copyright}). Finally, in the last section (\S \ref{sec:future_directions}), we argue for a comprehensive evaluation of creativity in AI that considers several dimensions of creativity and the creative process at its core. Furthermore, we discuss \textbf{future research directions} to enhance the creativity of AI systems, potentially drawing ideas from cognitive science and psychology.

Our goal in this survey is to provide a high-level yet comprehensive overview of the state of creativity in AI. 
We expect our survey to provide researchers working on machine creativity with comprehensive background knowledge and encourage them to explore new avenues for developing intelligent systems that can do creative generation.
\section{Creativity}
\label{sec:creativity}

\subsection{Defining Creativity}
While creativity as a concept seems intuitively easy to understand on the surface, there is still no consensus on what constitutes true creativity. This is primarily due to the subjective nature of creativity, as what is deemed novel and of quality can vary significantly across cultures, disciplines, and periods. \citet{aleinikov2000creating} lists more than $100$ proposed definitions, and the number keeps growing. Despite the lack of global consensus, there is one definition of creativity that has seen wide adoption by many philosophers and psychologists and has been dubbed as the {\textit{``standard definition''}} \cite{runco2012std, barron1955disposition, stein1953creativity}. According to this definition, creativity requires \textbf{\textit{novelty}} (a.k.a originality, uniqueness, etc.) and \textbf{\textit{value}} (a.k.a utility, effectiveness, usefulness, appropriateness, relevance, meaningfulness, etc.). 

The novelty criterion is typically self-explanatory to the point that people equate it to creativity in everyday life. However, many theorists have argued that novelty is insufficient for creativity, and value dimension is needed to filter out original nonsense, such as something generated by a truly random process. While value is generally understood as something inherently ``good'' for the respective audience, there appears to be such a thing as malevolent or ``dark'' creativity. For instance, one can be creative in producing torture instruments or in committing terrorist atrocities \cite{gaut2010}. Therefore, the interpretation of the value of a product as being ``effective'' towards its intended end, regardless of whether that end is morally good or bad, has been suggested as a better alternative \cite{Livingston2018-LIVEQ}. However, we should note that evaluating utility or value still requires an outside judgment which is subjective, can be faulty or biased and can change over time and across cultures. This is especially apparent in arts as there are many great artists (Bach, Van Gogh etc.) whose ``value'' have only been recognized longtime after their death. Moreover, sometimes novelty \textit{is} itself the value created by the artist because no one has done it before, particularly, in visual arts. Hence, some researchers have recently argued to drop the value criterion altogether from the definition of creativity \cite{brandt2021, weisberg2015}.

Despite its wide adoption, the sufficiency of the novelty and value conditions for creativity has also been challenged \cite{gaut2010, rudolph2001-cc, weisberg2015value}. It has been argued that \textbf{\textit{agency}}, a capacity to have beliefs, desires or intentional states, is a required attribute of creativity. For example, \citet{gaut2010} mentions the tectonic movement of the earth's crust that can produce valuable (financially and aesthetically) and sometimes original (new variation) diamonds, but we would hardly call tectonic movements creative. However, mere agency without \textbf{\textit{intentionality}} is also insufficient. \citet{gaut2010} illustrates this with an example where a person walking in a studio accidentally knocks over a set of paints, which spill onto a canvas and happen to create a beautiful and original painting. \citet{weisberg2015} has gone even further to suggest that creativity is simply \textit{intentional novelty}. 

While the creative process should be intentionally initiated, others have argued that the creative process should involve an element of \textbf{\textit{spontaneity}} \cite{Kronfeldner2009}. This allows the creative product to induce a \textit{surprise} in the audience since the output of the process is not foreseen from the beginning. Being ignorant of the end at the outset of the creative process opens the room for creativity as opposed to a mechanical routine or algorithm that, by definition, is exact and excludes any type of spontaneous modifications. To illustrate this contrast \citet{perkins2001eureka} makes a distinction between \textit{reasonable} problems (i.e. that can be reasoned out step-by-step such as anagrams) and \textit{unreasonable} problems (i.e. that are hard to describe with a step-by-step thinking).

The element of \textbf{\textit{surprise}} has been further developed by \cite{boden2004creative} into a widely recognized third dimension of the ``standard definition'' of creativity. This new definition can be seen as an elaborated version of the three criteria (i.e. new, useful and non-obvious) used by the United States Patent Office to determine whether an invention can come under patent protection\footnote{http://www.uspto.gov/inventors/patents.jsp} \cite{simonton2012}. In this survey, we will also take this extended definition as our working definition throughout the paper.

\subsection{Types of Creativity}
While creativity manifests itself in various forms across domains, even within a particular domain, different \textit{types} of creativity can be distinguished based on its timing or target audience and the difficulty level of the inherent creative process involved. A person might come up with a creative idea that is new to him/her but already invented by someone else in history. This is generally known as \textit{intrapersonal} or \textit{personal} (a.k.a psychological) creativity (often denoted as \textbf{P-creativity}), i.e. the product is novel within the frame of a person's life \cite{boden2004creative, stein1953creativity, weisberg1986creativity}. Researchers distinguish it from the \textit{interpersonal} or \textit{historical} creativity (often denoted as \textbf{H-creativity}), i.e. the product is novel with respect to the entire history of people such as Einstein's general relativity theory.

\textbf{Four-C} model of creativity, on the other hand, differentiates between four types of creativity corresponding to four levels of difficulty involved in producing creative artifacts \cite{kaufman2009fourc}. The first major type of creativity is known as \textbf{little-c} creativity which is what we find in everyday life as solutions to minor problems. Examples might include combining unusual ingredients to make a new type of meal or using a hand-held vacuum cleaner on the ceiling to remove flies. Almost everyone possesses this type of creativity in one way or another. The second main type of creativity in this model is the \textbf{Big-C} creativity that includes major works of scientific, technological, social, or artistic importance. Examples could be Darwin's theory of evolution, the invention of the printing press, or Leonardo Da Vinci's painting of the Mona Lisa. In addition to these two major categories, \citet{kaufman2009fourc} also defines two minor categories of creativity. First is the \textbf{mini-c} creativity for very small-scale cases of creativity such as young children's drawing or their creative experiments with Lego pieces. Second is the \textbf{Pro-C} creativity which is proposed for work produced by professional but non-prominent practitioners such as professional musicians or artists who generate novel work, but do not make historical contributions. 

Recently, there has been a suggestion to differentiate three types of creativity corresponding to three major levels of innovation that can be achieved \cite{demis-creativity2018}. First can be achieved through \textbf{interpolation} where a prototypical creative artifact is produced by averaging all the artifacts of the same class seen before. While it is an original product that did not exist before, it still relies heavily on the other existing products. An example would be to come up with a novel winning strategy in chess that is a combination of existing different strategies. Consequently, a second type of innovation can be achieved through \textbf{extrapolation} where a creative artifact extends the boundaries of what has been seen before, but is still of the same class. A completely new chess move that is not related to any existing moves can be seen as an example of extrapolative creativity. Finally, the highest level of creativity can be termed as \textbf{invention} where a creative artifact introduces a novel class of its own. Inventing chess itself or any major scientific invention is a perfect example of this type of creativity. This type of creativity typically requires \textit{transformation} of the existing conceptual space and is also known as \textbf{transformational} creativity \cite{boden2004creative}.

\subsection{Evaluation}
Evaluating creativity remains a challenging task in artificial intelligence due to its inherently subjective nature \cite{lamb2018eval}. Interestingly, some research work even argued against the quantitative evaluation of creativity, suggesting it is either too domain-specific to be measured effectively \cite{Baer2012DomainSA}, or that creativity is an inherently human trait that cannot be accurately modeled computationally \cite{Boden1991TheCM, Minsky1982WhyPT}. 
However, an overwhelming majority of the scientific community favors the possibility of computational modeling and evaluation of creativity \cite{veale2019computational}. Hence, numerous evaluation methods have been proposed in the past \cite{lamb2018eval}. However, most of the proposed metrics are either formal frameworks that are hard to implement in practice or manual psychometric creativity tests that require costly human involvement \cite{Kim2006CanWT} or automated metrics that are too domain-specific \cite{Frana2016RegentDependentCA}. We refer the reader to \citet{Franceschelli2021CreativityAM} and \citet{lamb2018eval} for more details on formal evaluation frameworks, and here we briefly summarize some of the relevant manual and automated metrics for creativity.

\subsubsection{Manual Evaluation}
Since creative products vary greatly in their forms and are hard to characterize with objective measures, the simplest and most common way to evaluate them is to ask other humans to manually rate them based on some criteria associated with creativity, which differs from task to task \cite{Lamb2015HumanCI}. For example, in story generation, humans are typically asked to rate a generated story on aspects such as \textbf{interestingness}, \textbf{coherence}, \textbf{relevance}, \textbf{humanlikeness} and etc. \cite{yang2024makes, goldfarb2020content, Rashkin2020PlotMachinesOG, yang-etal-2022-re3}. In other tasks where the goal is to produce multiple responses such as common psychometric creativity tests \textit{Alternative Uses Task (AUT)} \cite{guilford1967nature} and \textit{Torrance Tests of Creative Thinking (TTCT)} \cite{torrance1974torrance}, evaluation is centered around four dimensions of creativity: \textbf{\textit{fluency}} (the total number of meaningful, and relevant ideas generated in response to the stimulus), \textbf{\textit{flexibility}} (the number of different categories of relevant responses), \textbf{\textit{originality}} (the uniqueness or rarity of responses) and \textbf{\textit{elaboration}} (the amount of detail in the responses).

While it is common and straightforward to conduct human evaluation with ordinary humans, some have argued that people who are not experts on a kind of creative artifact might not be good judges of those artifacts \cite{Mirowski2022CoWritingSA, gervas2019, Veale2015GameOT, lamb2018eval, Lamb2015HumanCI}. This typically results in poor interrater reliability and even when they agree, their judgments do not correlate well with expert judgment \cite{lamb2018eval}. Therefore, it is generally recommended to employ \textbf{Consensual Assessment Technique} \cite{Amabile1983TheSP}, an evaluation method that relies on the collective judgment of experts in a given field. 

\subsubsection{Automated Evaluation}
While creativity is generally evaluated by humans, several attempts have also been made to devise automated measures of it \cite{organisciak2023beyond, Frana2016RegentDependentCA, cook2015generating, jordanous2015measuring, maher2012using}. These measures often target a specific dimension of creativity. Below, we review some automated measures for three dimensions of creativity: novelty, value, and surprise.
\paragraph{Novelty}
It is typically defined as the measure of how different an artifact is from other known artifacts in its class \cite{Maher2010EvaluatingCI}. Then a distance metric is established to quantify this difference based on the attributes of the artifact and the task space. For example, in the text generation task, a notion of \textbf{\textit{semantic distance}} is commonly employed as a distance measure \cite{Johnson2022DivergentSI, Beaty2020AutomatingCA, Prabhakaran2013ThinSO, Dunbar2009CreativityET, Harbison2014AutomatedSO}. More specifically, the text is embedded into a vector in semantic space and some distance or dissimilarity metric (e.g. typically \texttt{1-cosine\_similarity}) is used to compute how much semantically different is one text from another. However, the granularity of the text can differ from task to task. For example, in the story generation task, \citet{Karampiperis2014TowardsMF} defines the novelty of a story as the average semantic distance between the dominant terms included in the textual representation of the story, compared to the average semantic distance of the dominant terms in all stories where distance is measured based on the embeddings of terms. 

Novelty can also be characterized by the degree an artifact differs from the previously produced works that one has already seen \cite{Elgammal2015QuantifyingCI, Gunkle1975AestheticsAP}. This definition has inspired the development of Creative Adversarial Networks (CANs) \cite{elgammal2017can} similar to the popular Generative Adversarial Networks (GANs) \cite{goodfellow2014generative}. In CANs, the generator tries to fool the discriminator into thinking its generation is ``art'' and at the same time, the style of its generation is nothing known to the discriminator. Consequently, the score assigned by the discriminator (more specifically, \texttt{1-score}) can be used to measure the novelty of the generated artifact as suggested by \citet{Franceschelli2022DeepCreativityMC}.

\paragraph{Value}
This dimension is generally the hardest to evaluate as it depends on the subjective utility or performance of the artifact which is typically judged by domain experts of that artifact and can radically change across domains \cite{Maher2010EvaluatingCI}. In visual arts, this might correspond to ``beauty'', whereas in science to ``logical correctness''. Therefore, a metric appropriate for its domain should be employed. For example, in open-ended story generation, a minimally useful story can be defined as a relevant, coherent, and meaningful story. In this sense, automated metrics measuring the overall quality of a story can be leveraged \cite{xie-etal-2023-next, xie-etal-2023-deltascore, Guan2020UNIONAU, Chen2022StoryERAS}, however, it is often challenging to measure coherence \cite{zhao-etal-2023-discoscore, Laban2021CanTM}. Another more general evaluation of utility has been suggested by \citet{Franceschelli2022DeepCreativityMC} based on the discriminator score in GANs. Since in GANs, the discriminator learns the distribution of the real (and valuable) data, its score can directly be used as a proxy metric to measure value.

\paragraph{Surprise}
Also known as unexpectedness, surprise measures the artifact's degree of deviation from what is expected \cite{Maher2010EvaluatingCI}. Therefore, automatic metrics for surprise tend to be \textit{information-theoretic} \cite{kuznetsova-etal-2013-understanding, bunescu-uduehi-2022-distribution} and estimate the violation of expectation based on uncertainty reduction \cite{Frank2010UncertaintyRA, Hale2006UncertaintyAT}. However, semantic distance-based measures of surprise have also been suggested. For example, in the story generation task, \citet{Karampiperis2014TowardsMF} conceptualizes surprise as the average semantic distances between the consecutive fragments of a given story. Recent work has also suggested an automated measure based on the Bayesian theory of surprise \cite{Franceschelli2022DeepCreativityMC, Baldi2010OfBA}.
\definecolor{hidden-draw}{RGB}{205, 44, 36}
\definecolor{hidden-blue}{RGB}{194,232,247}
\definecolor{hidden-orange}{RGB}{243,202,120}
\definecolor{hidden-yellow}{RGB}{242,244,193}
\definecolor{hidden-green}{RGB}{0,255,127}
\definecolor{hidden-beige}{RGB}{237,232,208}
\definecolor{tree-level-1}{RGB}{245,20,85}
\definecolor{tree-level-2}{RGB}{246,86,118}
\definecolor{tree-level-3}{RGB}{248,177,193}
\definecolor{tree-leaf}{RGB}{176,230,198}

\tikzstyle{my-box}=[
    rectangle,
    draw=hidden-draw,
    rounded corners,
    text opacity=1,
    minimum height=1.5em,
    minimum width=5em,
    inner sep=2pt,
    align=center,
    fill opacity=.5,
]
\tikzstyle{lc-leaf}=[my-box, minimum height=1.5em,
    fill=hidden-orange!60, text=black, align=left,font=\scriptsize,
    inner xsep=2pt,
    inner ysep=4pt,
]
\tikzstyle{cps-leaf}=[my-box, minimum height=1.5em,
    fill=hidden-blue!60, text=black, align=left,font=\scriptsize,
    inner xsep=2pt,
    inner ysep=4pt,
]
\tikzstyle{ac-leaf}=[my-box, minimum height=1.5em,
    fill=hidden-beige!60, text=black, align=left,font=\scriptsize,
    inner xsep=2pt,
    inner ysep=4pt,
]
\tikzstyle{sc-leaf}=[my-box, minimum height=1.5em,
    fill=hidden-green!60, text=black, align=left,font=\scriptsize,
    inner xsep=2pt,
    inner ysep=4pt,
]

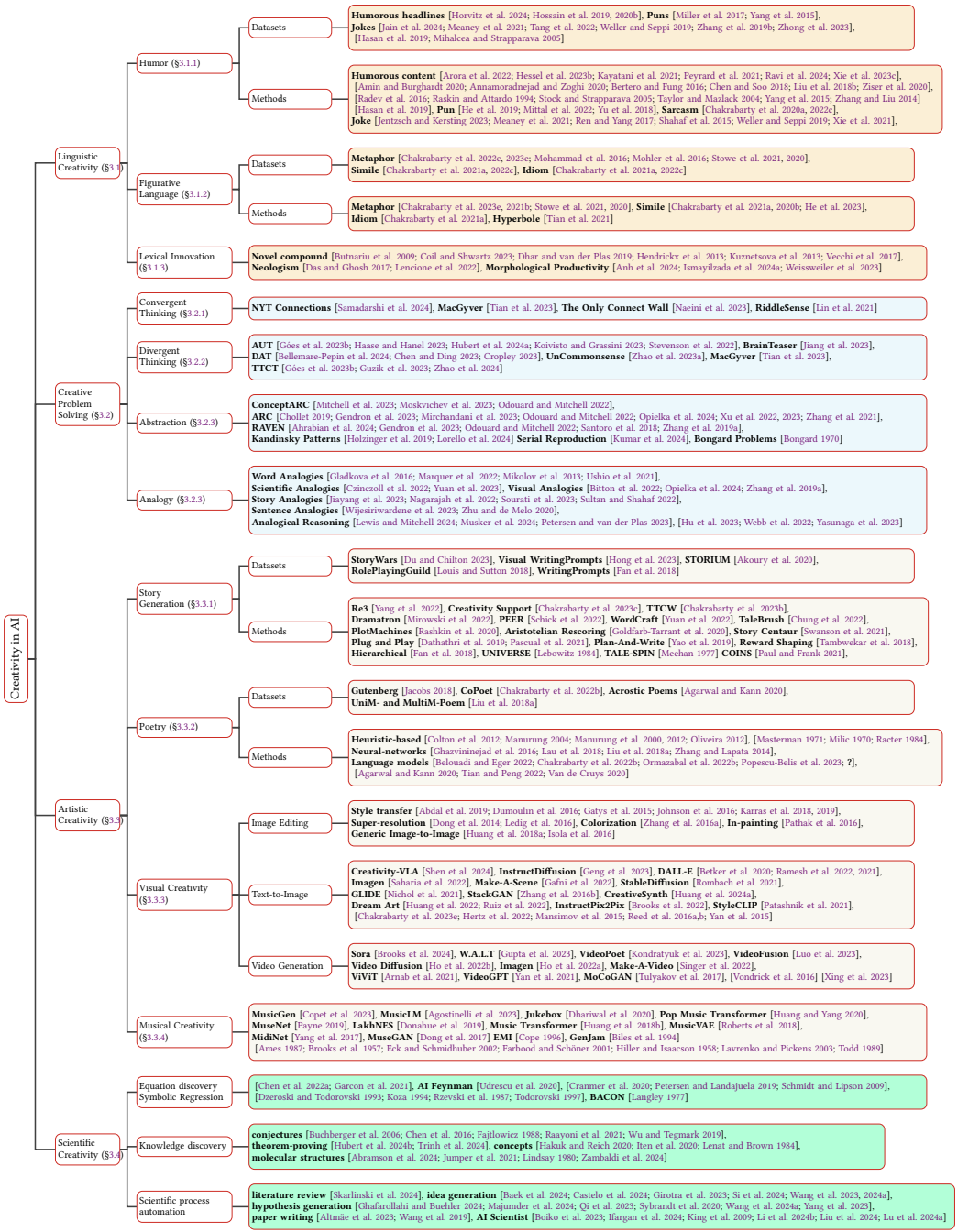
\begin{figure*}[tp]
    \centering
    \resizebox{\textwidth}{!}{
        \begin{forest}
            forked edges,
            for tree={
                grow=east,
                reversed=true,
                anchor=base west,
                parent anchor=east,
                child anchor=west,
                base=left,
                font=\small,
                rectangle,
                draw=hidden-draw,
                rounded corners,
                align=left,
                minimum width=4em,
                edge+={darkgray, line width=1pt},
                s sep=10pt,
                inner xsep=2pt,
                inner ysep=3pt,
                ver/.style={rotate=90, child anchor=north, parent anchor=south, anchor=center},
            },
            where level=1{text width=4.2em,font=\scriptsize,}{},
            where level=2{text width=6.4em,font=\scriptsize,}{},
            where level=3{text width=5.5em,font=\scriptsize,}{},
            where level=4{text width=6.1em,font=\scriptsize,}{},
        [
            Creativity in AI, ver
            [
                Linguistic \\ Creativity (\S \ref{sec:linguistic_creativity})
                [
                    Humor (\S \ref{sec:humor})
                    [
                        Datasets
                        [
                            \textbf{Humorous headlines} ~\cite{Horvitz2024GettingSA, hossain-etal-2020-stimulating, hossain-etal-2019-president}{,}
                            \textbf{Puns} ~\cite{miller-etal-2017-semeval, Yang2015HumorRA}{,}\\
                            \textbf{Jokes} ~\cite{tang2022naughtyformertransformerunderstandsoffensive, meaney-etal-2021-semeval, Jain2024IsAF, Zhong2023LetsTO, zhang-etal-2019-telling, weller-seppi-2019-humor}{,}\\
                            ~\cite{hasan-etal-2019-ur, mihalcea-strapparava-2005-making}
                            , lc-leaf, text width=40em
                        ]
                    ]
                    [
                        Methods
                        [
                            \textbf{Humorous content} ~\cite{ravi-etal-2024-small, hessel2023androids, Xie2023FunQATS, arora-etal-2022-transfer, Peyrard2021LaughingHC, kayatani2021}{,}\\
                            ~\cite{Annamoradnejad2020ColBERTUB, ziser2020, amin-burghardt-2020-survey, Chen2018HumorRU, liu-etal-2018-exploiting, bertero-fung-2016-deep}{,}\\
                            ~\cite{radev-etal-2016-humor, Yang2015HumorRA, zhang2014recognize, stock-strapparava-2005-hahacronym, taylor2004computationally, Raskin1994NonliteralnessAN}\\
                            ~\cite{hasan-etal-2019-ur}{,}
                            \textbf{Pun} ~\cite{mittal-etal-2022-ambipun, he-etal-2019-pun, Yu2018ANA}{,}
                            \textbf{Sarcasm} ~\cite{Chakrabarty2022FLUTEFL, chakrabarty-etal-2020-r}{,}\\
                            \textbf{Joke} ~\cite{jentzsch-kersting-2023-chatgpt, xie-etal-2021-uncertainty, meaney-etal-2021-semeval, weller-seppi-2019-humor, Ren2017NeuralJG, Shahaf2015InsideJI}{,}
                            , lc-leaf, text width=42em
                        ]
                    ]
                ]
                [
                    Figurative \\ Language (\S \ref{sec:figurative_lang})
                    [
                        Datasets
                        [
                            \textbf{Metaphor} ~\cite{Chakrabarty2023ISA, Chakrabarty2022FLUTEFL, Stowe2021MetaphorGW, Stowe2020MetaphoricPG, Mohammad2016MetaphorAA, mohler-etal-2016-introducing}{,}\\
                            \textbf{Simile} ~\cite{Chakrabarty2021ItsNR, Chakrabarty2022FLUTEFL}{,}
                            \textbf{Idiom} ~\cite{Chakrabarty2021ItsNR, Chakrabarty2022FLUTEFL}
                            , lc-leaf, text width=40em
                        ]
                    ]
                    [
                        Methods
                        [
                            \textbf{Metaphor} ~\cite{Chakrabarty2023ISA, chakrabarty-etal-2021-mermaid, Stowe2021MetaphorGW, Stowe2020MetaphoricPG}{,}
                            \textbf{Simile} ~\cite{he-etal-2023-hauser, Chakrabarty2021ItsNR, chakrabarty-etal-2020-generating}{,}\\
                            \textbf{Idiom} ~\cite{Chakrabarty2021ItsNR}{,}
                            \textbf{Hyperbole} ~\cite{Tian2021HypoGenHG}
                            , lc-leaf, text width=40em
                        ]
                    ]
                ]
                [
                    Lexical Innovation \\ (\S \ref{sec:novel_concepts})
                    [
                        \textbf{Novel compound} ~\cite{coil-shwartz-2023-chocolate, dhar2019learning, vecchi2017, kuznetsova-etal-2013-understanding, hendrickx-etal-2013-semeval, butnariu-etal-2009-semeval}{,}\\
                        \textbf{Neologism} ~\cite{lencione2022nameling, das-ghosh-2017-neuramanteau}{,}
                        \textbf{Morphological Productivity} ~\cite{ismayilzada2024evaluatingmorphologicalcompositionalgeneralization, anh-etal-2024-morphology, weissweiler-etal-2023-counting}
                        , lc-leaf, text width=48em
                    ]
                ]
            ]
            [
                Creative \\ Problem \\ Solving (\S \ref{sec:problem_solving})
                [
                    Convergent \\ Thinking (\S \ref{sec:convergent_think})
                    [
                        \textbf{NYT Connections} ~\cite{samadarshi2024connectingdotsevaluatingabstract}{,}
                        \textbf{MacGyver} ~\cite{Tian2023MacGyverAL}{,}
                        \textbf{The Only Connect Wall} ~\cite{naeini2023large}{,}
                        \textbf{RiddleSense} ~\cite{Lin2021RiddleSenseRA}
                        , cps-leaf, text width=48em
                    ]
                ]
                [
                    Divergent \\ Thinking (\S \ref{sec:divergent_think})
                    [
                        \textbf{AUT} ~\cite{Hubert2024TheCS, Koivisto2023BestHS, haase2023artificial, GesPushingGC, stevenson2022putting}{,}
                        \textbf{BrainTeaser} ~\cite{Jiang2023BRAINTEASERLT}{,}\\
                        \textbf{DAT} ~\cite{bellemare2024divergent, Chen2023ProbingTC, cropley2023artificial}{,}
                        \textbf{UnCommonsense} ~\cite{Zhao2023UNcommonsenseRA}{,}
                        \textbf{MacGyver} ~\cite{Tian2023MacGyverAL}{,}\\
                        \textbf{TTCT} \cite{Zhao2024AssessingAU, Guzik2023TheOO, GesPushingGC}
                        , cps-leaf, text width=48em
                    ]
                ]
                [
                    Abstraction (\S \ref{sec:abstraction})
                    [
                        \textbf{ConceptARC} ~\cite{Mitchell2023ComparingHG, Moskvichev2023TheCB, Odouard2022EvaluatingUO}{,}\\
                        \textbf{ARC} ~\cite{Opielka2024DoLL, Xu2023LLMsAT, Gendron2023LargeLM, Mirchandani2023LargeLM, Odouard2022EvaluatingUO, Xu2022GraphsCA, Zhang2021ACREAC, Chollet2019OnTM}{,}\\
                        \textbf{RAVEN} ~\cite{Ahrabian2024TheCC, Gendron2023LargeLM, Odouard2022EvaluatingUO, Zhang2019RAVENAD, Santoro2018MeasuringAR}{,}\\
                        \textbf{Kandinsky Patterns} ~\cite{Lorello2024TheKB, Holzinger2019KANDINSKYPA}
                        \textbf{Serial Reproduction} ~\cite{Kumar2024ComparingAI}{,}
                        \textbf{Bongard Problems} ~\cite{Bongard1970PatternR}
                        , cps-leaf, text width=48em
                    ]
                ]
                [
                    Analogy (\S \ref{sec:analogy})
                    [
                        \textbf{Word Analogies} ~\cite{Marquer2022TransferringLM, ushio-etal-2021-bert, gladkova-etal-2016-analogy, NIPS2013_9aa42b31}{,}\\
                        \textbf{Scientific Analogies} ~\cite{Yuan2023BeneathSS, czinczoll-etal-2022-scientific}{,}
                        \textbf{Visual Analogies} ~\cite{Opielka2024DoLL, bitton2022vasrvisualanalogiessituation, Zhang2019RAVENAD}{,}\\
                        \textbf{Story Analogies} ~\cite{jiayang-etal-2023-storyanalogy, Sourati2023ARNAR, sultan-shahaf-2022-life, nagarajah2022understandingnarrativesdimensionsanalogy}{,}\\
                        \textbf{Sentence Analogies} ~\cite{Wijesiriwardene2023ANALOGICALA, zhu-de-melo-2020-sentence}{,}\\
                        \textbf{Analogical Reasoning} ~\cite{Musker2024SemanticSI, Lewis2024UsingCT, Petersen2023CanLM}{,}
                        ~\cite{Yasunaga2023LargeLM, Hu2023InContextAR, Webb2022EmergentAR}
                        , cps-leaf, text width=48em
                    ]
                ]
            ]
            [
                Artistic \\ Creativity (\S \ref{sec:artistic-creativity})
                [
                    Story \\ Generation (\S \ref{sec:story_gen})
                    [
                        Datasets
                        [
                            \textbf{StoryWars} ~\cite{du-chilton-2023-storywars}{,}
                            \textbf{Visual WritingPrompts} ~\cite{hong-etal-2023-visual-writing}{,}
                            \textbf{STORIUM} ~\cite{akoury-etal-2020-storium}{,}\\
                            \textbf{RolePlayingGuild} ~\cite{louis-sutton-2018-deep}{,}
                            \textbf{WritingPrompts} ~\cite{Fan2018HierarchicalNS}
                            , ac-leaf, text width=40em
                        ]
                    ]
                    [
                        Methods
                        [
                            \textbf{Re3} ~\cite{yang-etal-2022-re3}{,}
                            \textbf{Creativity Support} ~\cite{Chakrabarty2023CreativitySI}{,}
                            \textbf{TTCW} ~\cite{chakrabarty2023art}{,}\\
                            \textbf{Dramatron} ~\cite{Mirowski2022CoWritingSA}{,}
                            \textbf{PEER} ~\cite{Schick2022PEERAC}{,}
                            \textbf{WordCraft} ~\cite{Yuan2022WordcraftSW}{,}
                            \textbf{TaleBrush} ~\cite{Chung2022TaleBrushSS}{,}\\
                            \textbf{PlotMachines} ~\cite{Rashkin2020PlotMachinesOG}{,}
                            \textbf{Aristotelian Rescoring} ~\cite{goldfarb2020content}{,}
                            \textbf{Story Centaur} ~\cite{Swanson2021StoryCL}{,}\\
                            \textbf{Plug and Play} ~\cite{Pascual2021APM, Dathathri2019PlugAP}{,}
                            \textbf{Plan-And-Write} ~\cite{yao2019plan}{,}
                            \textbf{Reward Shaping} ~\cite{Tambwekar2018ControllableNS}{,}\\
                            \textbf{Hierarchical} ~\cite{Fan2018HierarchicalNS}{,}
                            \textbf{UNIVERSE} ~\cite{lebowitz1984creating}{,}
                            \textbf{TALE-SPIN} ~\cite{Meehan1977TALESPINAI} 
                            \textbf{COINS} ~\cite{paul-frank-2021-coins}{,}
                            , ac-leaf, text width=41em
                        ]
                    ]
                ]
                [
                    Poetry (\S \ref{sec:poetry})
                    [
                        Datasets
                        [
                            \textbf{Gutenberg} ~\cite{jacobs2018}{,}
                            \textbf{CoPoet} ~\cite{chakrabarty-etal-2022-help}{,}
                            \textbf{Acrostic Poems} ~\cite{agarwal2020acrosticpoemgeneration}{,}\\
                            \textbf{UniM- and MultiM-Poem} ~\cite{Liu2018beyond}
                            , ac-leaf, text width=40em
                        ]
                    ]
                    [
                        Methods
                        [
                            \textbf{Heuristic-based} ~\cite{Oliveira2012PoeTryMeA, Colton2012FullFACEPG, Manurung2012UsingGA, Manurung2004AnEA, manurung2000flexible}{,}
                            ~\cite{Racter1984ThePB, masterman1971, Milic1970ThePU}{,}\\
                            \textbf{Neural-networks} ~\cite{Liu2018beyond, Lau2018DeepspeareAJ, ghazvininejad-etal-2016-generating, Zhang2014ChinesePG}{,}\\
                            \textbf{Language models} ~\cite{PopescuBelis2023GPoeTAL, Belouadi2022ByGPT5ES, chakrabarty-etal-2022-help, ormazabal-etal-2022-poelm}{,}\\ ~\cite{tian-peng-2022-zero, van-de-cruys-2020-automatic, agarwal2020acrosticpoemgeneration}
                            , ac-leaf, text width=42em
                        ]
                    ]
                ]
                [
                    Visual Creativity \\ (\S \ref{sec:visual_creativity})
                    [
                        Image Editing
                        [
                            \textbf{Style transfer} \cite{Karras2019AnalyzingAI, Abdal2019Image2StyleGANHT, Karras2018ASG, Johnson2016PerceptualLF, Dumoulin2016ALR, Gatys2015ANA}{,} \\
                            \textbf{Super-resolution} ~\cite{Ledig2016PhotoRealisticSI, Dong2014LearningAD}{,} \textbf{Colorization} ~\cite{Zhang2016ColorfulIC}{,} \textbf{In-painting} ~\cite{Pathak2016ContextEF}{,} \\
                            \textbf{Generic Image-to-Image} ~\cite{Huang2018MultimodalUI, Isola2016ImagetoImageTW}
                            , ac-leaf, text width=40em
                        ]
                    ]
                    [
                        Text-to-Image
                        [
                            \textbf{Creativity-VLA} ~\cite{Shen2024EmpoweringVC}{,} \textbf{InstructDiffusion} ~\cite{Geng2023InstructDiffusionAG}{,} \textbf{DALL-E} ~\cite{BetkerImprovingIG, Ramesh2022HierarchicalTI, Ramesh2021ZeroShotTG}{,} \\
                            \textbf{Imagen} ~\cite{Saharia2022PhotorealisticTD}{,} \textbf{Make-A-Scene} ~\cite{Gafni2022MakeASceneST}{,} \textbf{StableDiffusion} ~\cite{Rombach2021HighResolutionIS}{,} \\
                            \textbf{GLIDE} ~\cite{Nichol2021GLIDETP}{,} \textbf{StackGAN} \cite{Zhang2016StackGANTT}{,} \textbf{CreativeSynth} ~\cite{Huang2024CreativeSynthCB}{,} \\
                            \textbf{Dream Art} ~\cite{Ruiz2022DreamBoothFT, Huang2022DrawYA}{,} \textbf{InstructPix2Pix} ~\cite{Brooks2022InstructPix2PixLT}{,} \textbf{StyleCLIP} ~\cite{Patashnik2021StyleCLIPTM}{,}\\  ~\cite{Chakrabarty2023ISA, Hertz2022PrompttoPromptIE, Reed2016GenerativeAT, Reed2016LearningWA, Mansimov2015GeneratingIF, Yan2015Attribute2ImageCI}
                            , ac-leaf, text width=40em
                        ]
                    ]
                    [
                        Video Generation
                        [
                            \textbf{Sora} ~\cite{videoworldsimulators2024}{,} \textbf{W.A.L.T} ~\cite{Gupta2023PhotorealisticVG}{,} \textbf{VideoPoet} ~\cite{Kondratyuk2023VideoPoetAL}{,}
                            \textbf{VideoFusion} ~\cite{Luo2023VideoFusionDD}{,} \\
                            \textbf{Video Diffusion} ~\cite{Ho2022VideoDM}{,} \textbf{Imagen} ~\cite{Ho2022ImagenVH}{,}
                            \textbf{Make-A-Video} ~\cite{Singer2022MakeAVideoTG}{,} \\\textbf{ViViT} ~\cite{Arnab2021ViViTAV}{,} \textbf{VideoGPT} ~\cite{Yan2021VideoGPTVG}{,}
                            \textbf{MoCoGAN} ~\cite{Tulyakov2017MoCoGANDM}{,} ~\cite{Vondrick2016GeneratingVW} ~\cite{Xing2023ASO}
                            , ac-leaf, text width=40em
                        ]
                    ]
                ]
                [
                    Musical Creativity \\ (\S \ref{sec:musical_creativity})
                    [
                        \textbf{MusicGen} ~\cite{Copet2023SimpleAC}{,} \textbf{MusicLM} ~\cite{Agostinelli2023MusicLMGM}{,} \textbf{Jukebox} ~\cite{Dhariwal2020JukeboxAG}{,}
                        \textbf{Pop Music Transformer} ~\cite{Huang2020PopMT}{,} \\
                        \textbf{MuseNet} ~\cite{payne2019musenet}{,} \textbf{LakhNES} ~\cite{Donahue2019LakhNESIM}{,}
                        \textbf{Music Transformer} ~\cite{Huang2018MusicTG}{,} \textbf{MusicVAE} ~\cite{Roberts2018AHL}{,} \\
                        \textbf{MidiNet} ~\cite{Yang2017MidiNetAC}{,}
                        \textbf{MuseGAN} ~\cite{Dong2017MuseGANMS} \textbf{EMI} ~\cite{cope1996experiments}{,} \textbf{GenJam} ~\cite{biles1994genjam} \\
                        ~\cite{lavrenko2003polyphonic, eck2002finding, farbood2001analysis, todd1989connectionist, ames1987automated, brooks1957experiment, hiller1953musical}
                         , ac-leaf, text width=48em
                    ]
                ]
            ]
            [
                Scientific \\ Creativity (\S \ref{sec:scientific-creativity})
                [
                    Equation discovery \\ Symbolic Regression
                    [
                        ~\cite{Chen2022AutomatedDO, Garcon2021DeepNN}{,} \textbf{AI Feynman} ~\cite{Udrescu2020AIF2}{,}
                        ~\cite{Cranmer2020DiscoveringSM, Petersen2019DeepSR, schmidt2009distilling}{,} \\
                        ~\cite{CoTodorovski1997DeclarativeBI, Koza1994GeneticPA, Deroski1993DiscoveringD, Rzevski1987ScientificDC}{,} \textbf{BACON} ~\cite{Langley1977BACONAP}
                         , sc-leaf, text width=48em
                    ]
                ]
                [
                   Knowledge discovery
                   [
                        \textbf{conjectures} ~\cite{Raayoni_2021, wu2019toward, chen2016automated, BUCHBERGER2006470, FAJTLOWICZ1988113}{,} \\
                        \textbf{theorem-proving} ~\cite{hubert2024, Trinh2024}{,} \textbf{concepts} ~\cite{HAKUK2020101080, PhysRevLett2020, lenat1984and}{,} \\
                        \textbf{molecular structures} ~\cite{Abramson2024, zambaldi2024novodesignhighaffinityprotein, Jumper2021, lindsay1980applications}
                         , sc-leaf, text width=45em
                   ]
                ]
                [
                    Scientific process \\ automation
                    [
                        \textbf{literature review} ~\cite{skarlinski2024language}{,} \textbf{idea generation} ~\cite{si2024llmsgeneratenovelresearch, Baek2024ResearchAgentIR, castelo2024ai, wang2024scimonscientificinspirationmachines, Girotra2023IdeasAD, Wang2023SciMONSI}{,} \\
                        \textbf{hypothesis generation} ~\cite{wang2024scimonscientificinspirationmachines, Ghafarollahi2024SciAgentsAS, majumder2024discoverybenchdatadrivendiscoverylarge, qi2023largelanguagemodelszero, Yang2023LargeLM, Sybrandt2020}{,} \\
                        \textbf{paper writing} ~\cite{altmae2023, Wang2019PaperRobotID}{,} 
                        \textbf{AI Scientist} ~\cite{lu2024aiscientistfullyautomated, ifargan2024autonomousllmdrivenresearchdata, Li2024MLRCopilotAM, liu2024coquest, boiko2023emergentautonomousscientificresearch, king2009automate}
                          , sc-leaf, text width=50em
                    ]
                ]
            ]
        ]
        \end{forest}
    }
    \caption{Taxonomy of creativity in AI covering areas of linguistic creativity, creative problem-solving, artistic and scientific creativity. Note that this taxonomy is not exhaustive, but rather a representative view of the key works.}
    \label{fig:creativity_taxonomy}
\end{figure*}

\section{Domains of Creativity}
\label{sec:domains}
Creativity is a multifaceted concept that spans across various domains, each harnessing its unique form of imaginative thought and innovation. In this section, we will review the state of creativity in AI across four major domains where machine creativity is most extensively explored: linguistics, art, science, and problem-solving. 
\subsection{Linguistic Creativity}
\label{sec:linguistic_creativity}
The creative aspect of language in linguistics has been discussed since the early days \cite{chomsky1965aspects}. Chomsky, in this paper, attributes creativity mainly to the essential property of language to provide means to express many thoughts indefinitely. However, several linguists since Chomsky have argued against using this characterization since it does not align with the everyday definition of creativity \cite{bergs2019ling, sampson2017ling, zawada2006ling}. Chomsky's theory of grammar might generate an infinite number of sentences; it, however, relies on a fixed set of rules, while creativity requires deviation from rules. In this sense, \citet{sampson2017ling} suggests distinguishing between \textbf{F-creativity} (fixed) and \textbf{E-creativity} (extending), where F-creativity refers to the Chomskian interpretation of linguistic creativity (a.k.a productivity in morphology) and E-creativity corresponds to the real linguistic innovation such as metaphors, jokes, neologisms, etc. Some recent works have explored the F-creativity of large language models and found that this task is challenging in general and even harder in more morphological complex languages \cite{ismayilzada2024evaluatingmorphologicalcompositionalgeneralization, anh-etal-2024-morphology, weissweiler-etal-2023-counting}. Most past works however have focused on studying the E-creativity of AI systems which we review in the following sections.

\subsubsection{Humor}
\label{sec:humor}
Humor is one of the most common ways in which humans creatively use language to express their ideas and feelings. Early works to model humor focused on hand-crafted linguistic templates and wordplay \cite{stock-strapparava-2005-hahacronym, taylor2004computationally, Raskin1994NonliteralnessAN}. Subsequent works have leveraged language's lexical and syntactic properties as humor-specific features for humor detection \cite{liu-etal-2018-exploiting, Yang2015HumorRA, zhang2014recognize}. The growing interest in computational humor in recent years has resulted in several shared tasks organized by the NLP community \cite{ meaney-etal-2021-semeval, hossain-etal-2020-semeval, van-hee-etal-2018-semeval, Castro2018OverviewOT, potash-etal-2017-semeval, miller-etal-2017-semeval}. Latest works have developed methods based on neural networks and language models to generate and detect \textbf{humorous content} \cite{ravi-etal-2024-small, arora-etal-2022-transfer, Peyrard2021LaughingHC, Annamoradnejad2020ColBERTUB, ziser2020, amin-burghardt-2020-survey, hossain-etal-2019-president, Chen2018HumorRU, bertero-fung-2016-deep}, \textbf{jokes} \cite{Horvitz2024GettingSA, tang2022naughtyformertransformerunderstandsoffensive, xie-etal-2021-uncertainty, weller-seppi-2019-humor, Ren2017NeuralJG}, \textbf{puns} \cite{he-etal-2019-pun, mittal-etal-2022-ambipun, Yu2018ANA}, and \textbf{sarcasm} \cite{Chakrabarty2022FLUTEFL, chakrabarty-etal-2020-r}. Several datasets have also been proposed to benchmark the humor capacity of LLMs in several languages including English \cite{Horvitz2024GettingSA, Jain2024IsAF, tang2022naughtyformertransformerunderstandsoffensive, meaney-etal-2021-semeval, hossain-etal-2020-stimulating, hossain-etal-2019-president, miller-etal-2017-semeval}, Chinese \cite{zhang-etal-2019-telling}, Italian \cite{buscaldi2007some}, Spanish \cite{castro2017crowd}, Dutch \cite{winters2020dutch} and Russian \cite{blinov2019large}. Computational humor has also been explored in multimodal settings involving images, audio, and video in addition to text \cite{hessel-etal-2023-androids, Xie2023FunQATS, hasan-etal-2019-ur, bertero-fung-2016-deep, radev-etal-2016-humor, Shahaf2015InsideJI}. While the latest methods particularly LLMs show an impressive ability to generate and detect humorous content, recent work has also shown that these models still fail to \textit{reliably understand} humor \cite{hessel-etal-2023-androids, GesIsGG2023, kocon2023chatgpt, borji2023categoricalarchivechatgptfailures} and generated jokes typically \textit{lack diversity} \cite{jentzsch-kersting-2023-chatgpt} which has been attributed to training on less diverse humor datasets \cite{baranov-etal-2023-told}. Creative training frameworks have also been developed to improve the humor generation capabilities of LLMs \cite{Zhong2023LetsTO}. We refer the reader to \citet{amin-burghardt-2020-survey} for an in-depth survey on computational humor.

\subsubsection{Figurative Language}
\label{sec:figurative_lang}
Figurative language is a term in language studies encompassing various figures of speech like hyperbole, similes and metaphor \cite{veale2016metaphor, roberts1994people, Paul1970-PAUFL}. These elements can be used to achieve a range of communicative goals.
Figurative language generation involves transforming a text into a specific figure of speech while maintaining the original meaning \citep{Lai2024ASO}. Generating figurative language requires an understanding of abstract concepts, commonsense reasoning, and an ability to make analogies and deviate from literal meaning. Recent works have shown that language models with injected commonsense knowledge can generate textual and visual \textbf{metaphors} \cite{chakrabarty-etal-2023-spy, chakrabarty-etal-2021-mermaid}, \textbf{similes} \cite{chakrabarty-etal-2020-generating, he-etal-2023-hauser}, \textbf{idioms} \cite{Chakrabarty2021ItsNR} and \textbf{hyperboles} \cite{Tian2021HypoGenHG}. \citet{Chakrabarty2023ArtOA} reveals that metaphors generated by large language models are often \textit{incoherent or cliched}. \citet{Chakrabarty2023ArtOA} highlights the following example of such a metaphor generated by an LLM: \textit{``- However, she managed to laugh louder and louder until her laughter transformed into an embrace of the sun’s atmosphere.''} We refer the reader to \citet{Lai2024ASO} and \citet{abulaish2020} for an in-depth survey on the automatic generation and detection of figurative language.

\begin{figure*}[t]
\includegraphics[width=\textwidth]{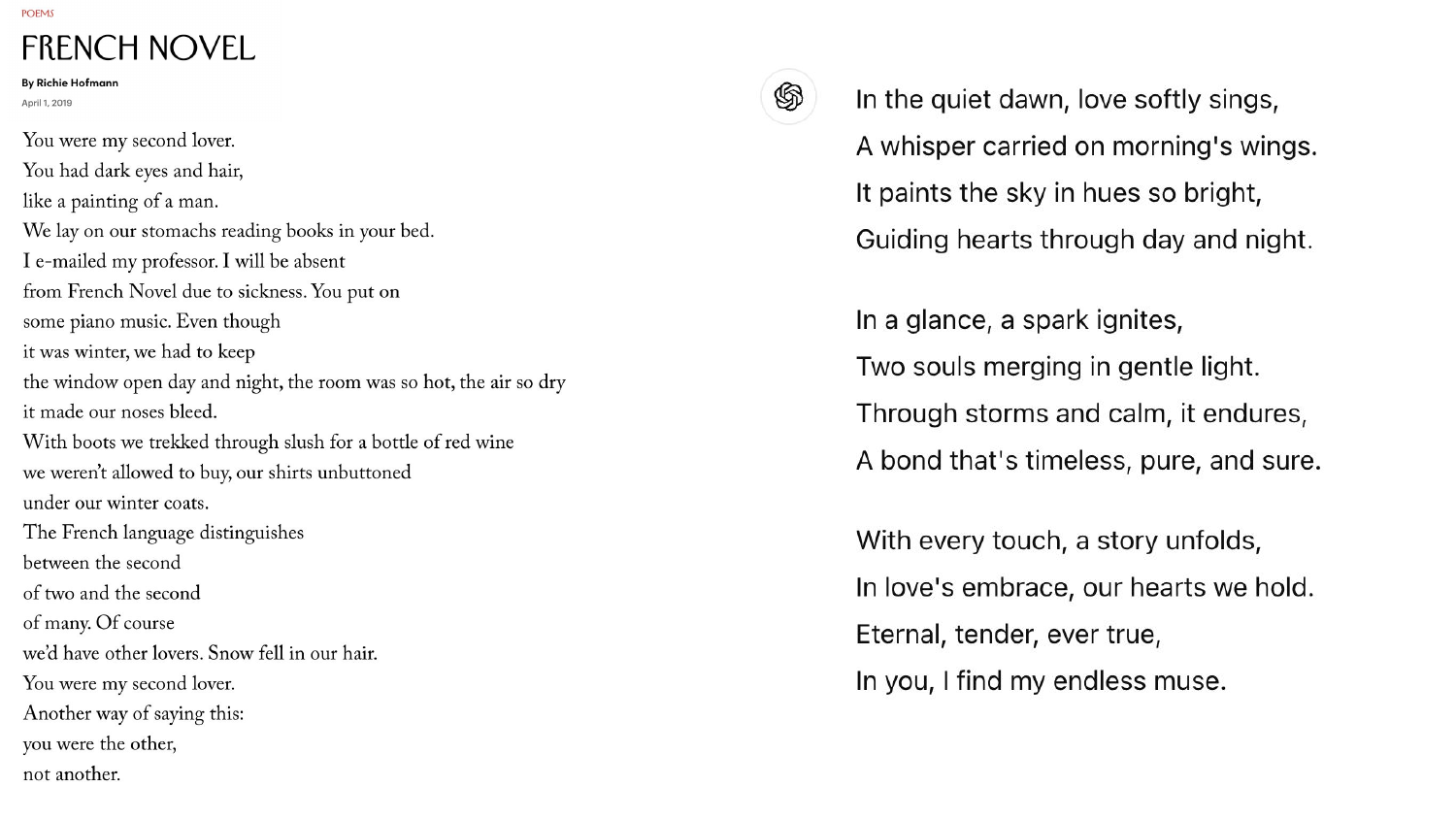}
\caption{Illustration of a qualitative difference between poetry written by humans and machines. \textbf{Left:} Poem about love published on New Yorker. \textbf{Right:} Poem about love generated by GPT-4o. While New Yorker poem draws deep metaphoric parallels between linguistic features of French and love, the GPT-4o generated poem merely describes love using cliché phrases. A similar comparison was made between Grok and the same New Yorker poem in \cite{chakrabarty-etal-2023-creative}.}
\label{fig:poetry-ex}
\end{figure*}

\subsubsection{Lexical Innovation}
\label{sec:novel_concepts}
Understanding and generating novel words or word compounds is a challenging linguistic task that often requires creativity, commonsense knowledge, and an ability to generalize over seen concepts \cite{costello2000efficient, wisniewski1997concepts}. Similar \textbf{noun compounds} might have different meanings based on our common understanding. For example, knowing that \textit{``chocolate croissant''} means a \textit{``croissant filled with chocolate''} does not necessarily imply that  \textit{``chocolate bunny''} would mean \textit{``a bunny filled with chocolate''}, but rather a piece of \textit{``chocolate in the shape of a bunny''}. Several works have evaluated and analyzed language models on the task of interpreting and predicting the emergence of these noun compounds and found that models generally show a moderate performance \cite{coil-shwartz-2023-chocolate, dhar2019learning, kuznetsova-etal-2013-understanding}. Other works have successfully trained neural networks to generate \textbf{neologisms} (i.e. newly coined words or phrases) \cite{das-ghosh-2017-neuramanteau, lencione2022nameling}. On the other hand, previous works have also shown that large language models can fail at \textit{linguistic generalization} tasks such as morphologically deriving new words from nonce roots \cite{ismayilzada2024evaluatingmorphologicalcompositionalgeneralization, weissweiler-etal-2023-counting} and can occasionally \textit{duplicate} large amounts of text from its training data \cite{McCoy2021HowMD}. Similarly, a recent work explores the linguistic creativity of both large language models and humans by reconstructing their text output from the existing text snippets on the web and finds that the seemingly remarkable creativity of model outputs may be in large part attributable to the remarkable creativity of human-written texts on the web \cite{lu2024aihumanityssalieriquantifying}.
\subsection{Creative Problem-Solving}
\label{sec:problem_solving}
Creative problem-solving is the mental process of searching and coming up with creative solutions to a given problem \cite{duncker1948problem}. It is a challenging task for machines as it not only requires creativity but also commonsense reasoning, and compositional generalization \cite{davidson_gureckis_lake_2022}. In addition, creatively solving a problem is usually characterized by two kinds of thinking, namely, \textbf{convergent} and \textbf{divergent} thinking, and involves deep \textbf{abstraction and analogy-making} abilities.

\begin{figure}[t]
\begin{subfigure}[c]{0.45\textwidth}
    \centering
    \includegraphics[width=\linewidth]{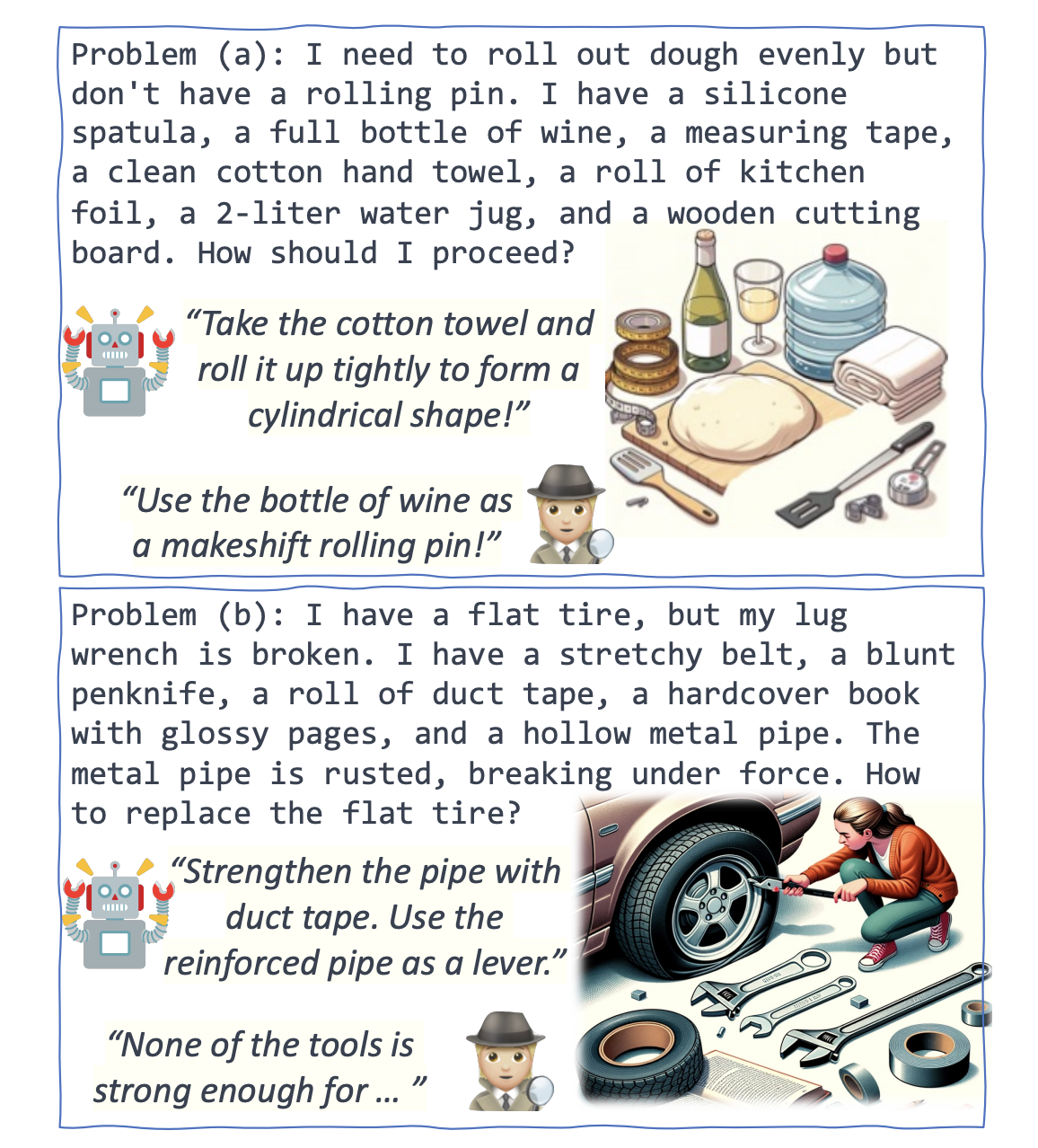}
    \caption{Example from the MacGyver dataset for creative problem-solving \cite{Tian2023MacGyverAL}. Problems in this dataset require innovative usage of objects and involve both convergent and divergent thinking.}
    \label{fig:macgyver-ex}
\end{subfigure}
\hfill
\begin{subfigure}[c]{0.45\textwidth}
    \centering
    \includegraphics[width=\linewidth]{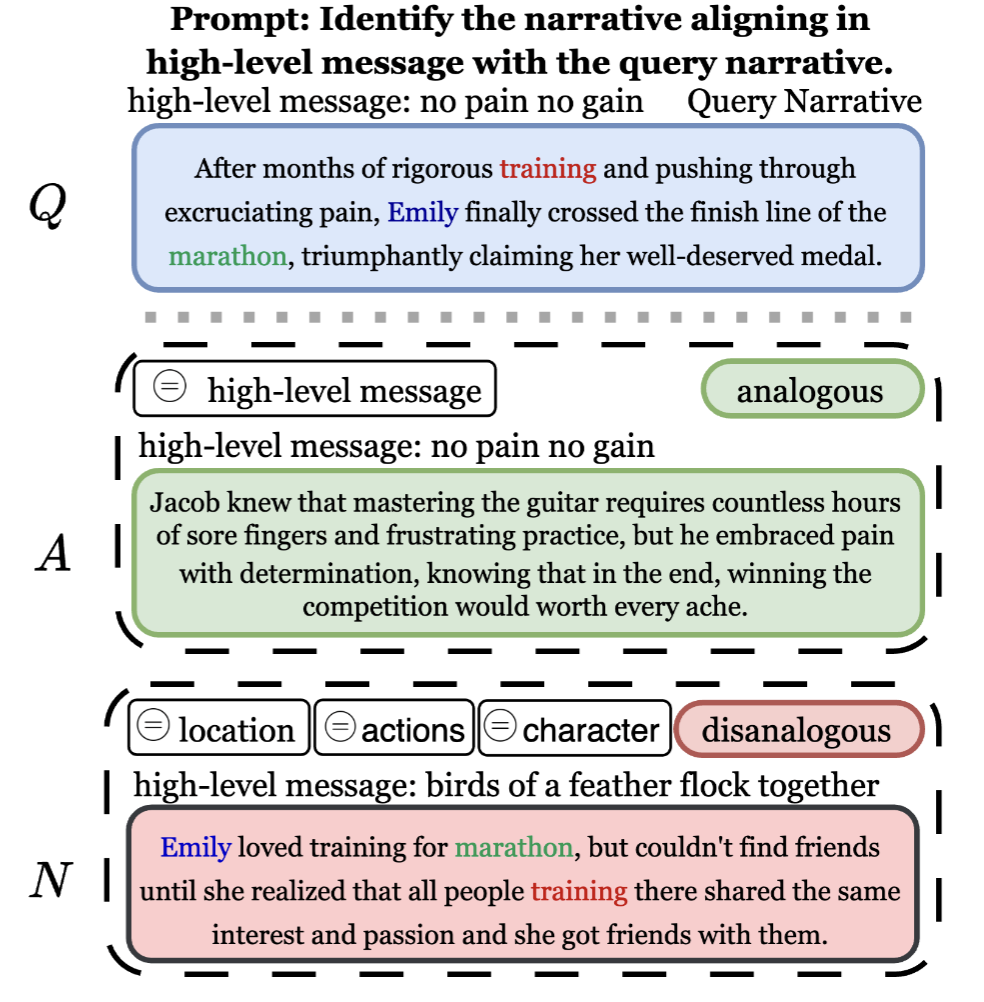}
    \caption{Example from Analogical Reasoning over Narratives benchmark \cite{Sourati2023ARNAR}. The task is to distinguish between analogous narrative $A$ and distractor $N$ for the query narrative $Q$.}
    \label{fig:analogy-ex}
\end{subfigure}
\caption{Examples from Creative Problem-Solving datasets.}
\label{fig:macgyver-analogy-ex}
\end{figure}

\subsubsection{Convergent Thinking}
\label{sec:convergent_think}
Convergent thinking models creativity in terms of an ability to produce a single optimal solution for a given problem \cite{guilford1967nature}. This type of creativity requires one to be able to associate seemingly remote ideas and converge to a unified solution. To evaluate this type of thinking in humans, psychologists have come up with several creativity tests such as \textbf{Remote Associates Test (RAT)} \cite{Mednick1962TheAB} and \textbf{insight problems} \cite{webb2017once}. For example, the goal in RAT is to connect several unrelated words with one concept, e.g. words \textit{``broken'', ``clear''} and \textit{``eye''} can be connected with the word \textit{``glass''}. 

Language models have recently been evaluated on problems that require convergent thinking. \citet{Lin2021RiddleSenseRA} tests language models on solving riddles that require creativity and commonsense and finds a \textit{significant gap} between model and human performance. \citet{naeini2023large} uses the popular British quiz show Only Connect's Connecting Wall that mimics RAT formulation with built-in, deliberate red herrings (i.e. misleading stimuli or distractors) and evaluates large language models such as GPT-4 on these problems. They report poor model performance and show that models are highly \textit{susceptible to distractors} in the input and manifest a form of \textbf{\textit{fixation effect}} (a.k.a functional fixedness or Einstellung effect) \cite{Wiley1998ExpertiseAM, Smith1991IncubationAT, Barber1960RigidityOB}. This type of cognitive bias forces the model to fixate on its past knowledge and prevents it from thinking ``out-of-the-box''. The same effect is also found when models are evaluated on everyday problems involving unconventional use of objects \cite{Tian2023MacGyverAL}. Very recently, large language models such as GPT-4o have been evaluated on the popular New York Times game \textit{Connections} and have been found to struggle with \textit{associating} encyclopedic and linguistic knowledge at an abstract level \cite{samadarshi2024connectingdotsevaluatingabstract}. Another study investigating both convergent and divergent creativity of language models has revealed that language models also fall short of demonstrating human-like convergent creativity in code generation \cite{Lu2024BenchmarkingLM}. 

\subsubsection{Divergent Thinking}
\label{sec:divergent_think}
Divergent thinking requires one to conceptualize multiple, often seemingly disconnected ideas \cite{guilford1967nature}. It essentially plays the opposite role to convergent thinking and therefore, the goal is to start with a unified idea and diverge from this idea into the space of all ideas to find the ones that are relevant to the task at hand. Psychologists have also devised creativity tests to evaluate humans' divergent thinking abilities, such as \textbf{Alternate Uses Test (AUT)} \cite{guilford1967nature} and \textbf{Torrance Tests of Creative Thinking (TTCT)}. AUT tests creativity based on whether the participant can come up with unusual (creative) uses for an everyday object and the results are typically evaluated either manually or using \textit{semantic distance}. For example, a \textit{``brick''} can be used as a
\textit{``paperweight''} or \textit{``to break a window''} and \textit{``coffee cup''} can be used as \textit{``small bowl''}, or \textit{``a hat for an elf''} etc. TTCT consists of several verbal and non-verbal tasks such as imagining impossibilities or the consequences of actions. Works evaluating GPT-3 \cite{brown2020language} and GPT-4 \cite{openai2023gpt4} on these tests report near-human performance results \cite{Hubert2024TheCS, Zhao2024AssessingAU, haase2023artificial, Guzik2023TheOO, Koivisto2023BestHS, GesPushingGC, stevenson2022putting}. Other tests that highly correlate with human creativity measured by AUT have also been proposed such as the task of \textbf{\textit{naming unrelated words}} (a.k.a Divergent Associations Task) \cite{Olson2021NamingUW}. Some recent works have used this test to evaluate the creativity of large language models and found that models outperform humans \cite{bellemare2024divergent, cropley2023artificial, Chen2023ProbingTC}.

While language models perform strongly on AUT-like divergent thinking tasks, they, however, struggle when these tasks require some form of \textbf{\textit{lateral thinking}} or \textbf{\textit{``thinking out-of-the-box''}} \cite{huang-etal-2024-lateval}. For example, recent works have found that defying default commonsense associations and modeling \textit{unexpected} or \textit{unlikely} situations are challenging for large language models \cite{Jiang2023BRAINTEASERLT, Tian2023MacGyverAL, Zhao2023UNcommonsenseRA}. Figure \ref{fig:macgyver-ex} illustrates a creative problem-solving example from \citet{Tian2023MacGyverAL} that involves unconventional use of everyday objects.

\subsubsection{Abstraction and Analogy-Making}
\label{sec:abstraction_analogy}
Conceptual abstraction and analogy-making lie at the core of human cognition and intelligence \cite{Chollet2019OnTM, Mitchell_2021, hofstadter2001}. These are abilities that enable humans to generalize to new domains, invent novel concepts, and make useful and often surprising connections between concepts. In other words, abstraction and analogy-making serve as foundational building blocks for creative thinking. 

\begin{figure*}[t]
\includegraphics[width=\textwidth]{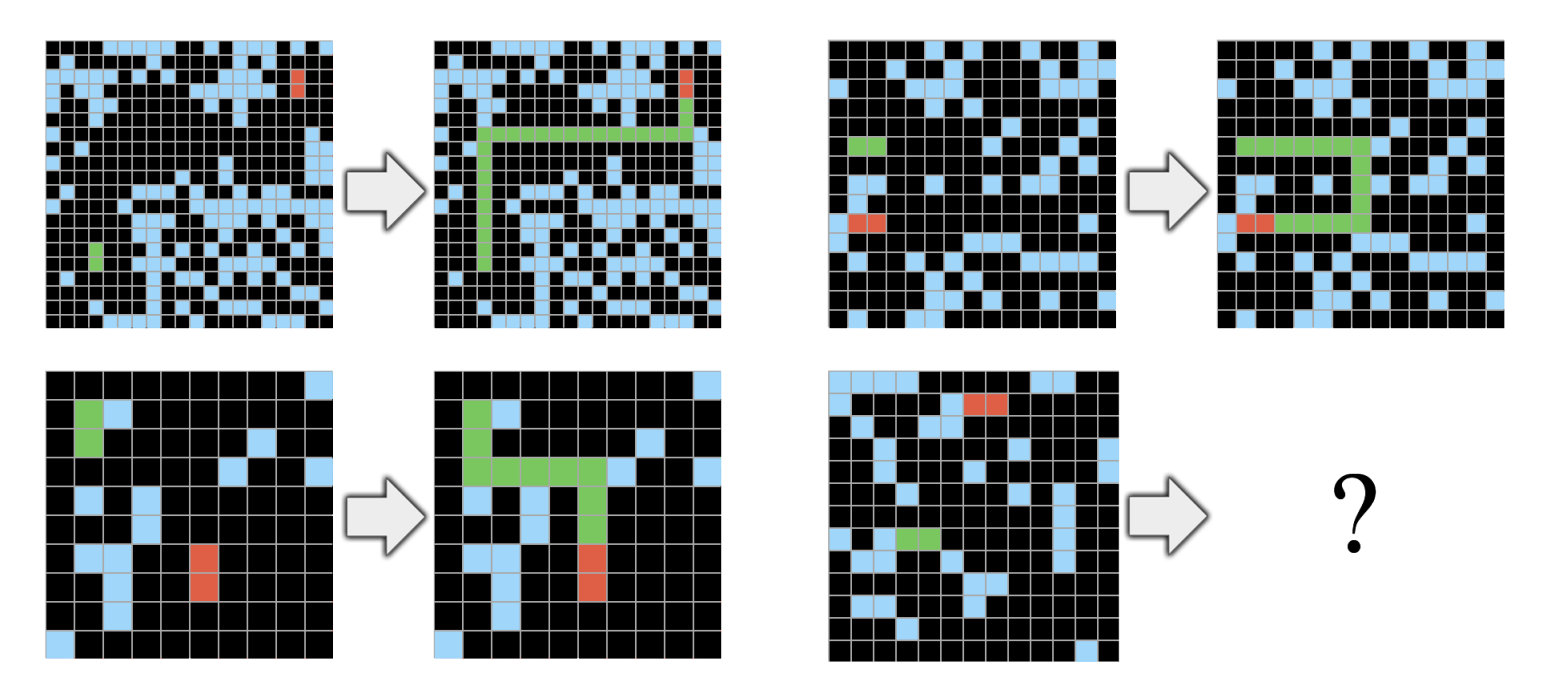}
\caption{Example from the Abstraction and Reasoning Corpus (ARC) \cite{Chollet2019OnTM} designed to test the abstractive thinking capabilities of both humans and machines.}
\label{fig:arc-ex}
\end{figure*}

\paragraph{Abstraction}
\label{sec:abstraction}
As the cornerstone of human intelligence, abstraction, and abstract reasoning are typically evaluated using visual IQ tests in humans. Popular examples of these tests are the \textbf{RAVEN progressive matrices} \cite{John2003RavenPM}, \textbf{Bongard problems} \cite{Bongard1970PatternR} and the recently introduced \textbf{Kandinsky Patterns} \cite{Holzinger2019KANDINSKYPA}, the \textbf{Abstraction and Reasoning Corpus (ARC)} \cite{Chollet2019OnTM} and its variations \cite{Moskvichev2023TheCB}. These tests require the participants to identify and complete an abstract visual pattern based on given examples. Although several attempts have been made to solve these tasks using both symbolic-based and neural network-driven approaches \cite{Lorello2024TheKB, Mirchandani2023LargeLM, Hu2023InContextAR, Xu2022GraphsCA, Santoro2018MeasuringAR}, modern AI systems still struggle with solving \textit{RAVEN-like} \cite{Ahrabian2024TheCC, Gendron2023LargeLM, Odouard2022EvaluatingUO, Zhang2019RAVENAD} and \textit{ARC-like} tasks \cite{Moskvichev2023TheCB, Mitchell2023ComparingHG, Xu2023LLMsAT, Odouard2022EvaluatingUO, Zhang2021ACREAC}. Analysis of abstraction via a \textbf{serial reproduction task} \cite{Langlois2021SerialRR} where participants are asked to produce a textual stimulus for the next participant upon observing a visual stimulus and vice versa, has suggested that GPT-4 unlike humans relies heavily on linguistic representations even in vision-only paradigm \cite{Kumar2024ComparingAI}. Figure \ref{fig:arc-ex} illustrates an example from the ARC task \cite{Chollet2019OnTM}. The problems in this corpus are quite hard to solve to the extent that this task has been recognized as the \textit{de facto} benchmark for measuring progress towards Artificial General Intelligence (AGI) and a public competition with a grand prize of \$$1,000,000$ has recently been launched\footnote{https://arcprize.org/}. At the time of writing this paper, the highest score is $49.5\%$ far from the passing threshold of $85\%$ (human-level).

\paragraph{Analogy-Making}
\label{sec:analogy}
In its basic form, analogy-making is the ability to identify a relation between two concepts and apply it to a new concept. For example, \textit{Paris} is to \textit{France} as \textit{Tokyo} is to \textit{Japan} (i.e. capital:country relation). Early approaches to computational analogy-making were symbolic-based and required extensive hand-coded input i.e. structured representations of both the entities and their relations \cite{Turney2008TheLR, Gentner1983StructureMappingAT, Falkenhainer1989TheSE}. Later, word embedding models based on neural networks were shown to exhibit analogy-making abilities at the word level and most works focused on a limited set of analogy types based on a handful of relations that are often of a morphological nature. \cite{Marquer2022TransferringLM, gladkova-etal-2016-analogy, NIPS2013_9aa42b31}.  These do not encompass the typical analogical reasoning humans perform in everyday life about complex situations. More recent work has focused on including a multitude of relations and datasets that are used to test analogical reasoning in humans. \cite{Petersen2023CanLM, ushio-etal-2021-bert, Jacob2023FAMEFS}

While some works have argued for \textit{emergent analogical reasoning} abilities of large language models \cite{Webb2022EmergentAR, Yasunaga2023LargeLM, Hu2023InContextAR}, other works have shown that these models lack the \textit{robustness and generality} exhibited by humans when it comes to \textbf{long text analogies} \cite{Wijesiriwardene2023ANALOGICALA, zhu-de-melo-2020-sentence}, \textbf{scientific analogies} \cite{Yuan2023BeneathSS, czinczoll-etal-2022-scientific}, \textbf{story analogies} \cite{jiayang-etal-2023-storyanalogy, Sourati2023ARNAR, sultan-shahaf-2022-life, nagarajah2022understandingnarrativesdimensionsanalogy}, \textbf{visual analogies} \cite{Opielka2024DoLL, bitton2022vasrvisualanalogiessituation, Zhang2019RAVENAD} and \textbf{complex analogical reasoning} \cite{Musker2024SemanticSI, Lewis2024UsingCT}. Figure \ref{fig:analogy-ex} illustrates an example from the analogical reasoning over narratives benchmark \cite{Sourati2023ARNAR}.
\subsection{Artistic Creativity}
\label{sec:artistic-creativity}
Artistic creativity is the ability to produce original, imaginative, and expressive works in various art forms, such as creative writing, poetry, visual arts, music, dance, theater, and more. In this section, we will focus on the advancements made in AI to produce creative stories, poetry, visual, and musical content automatically and also point out the remaining challenges.

\subsubsection{Story Generation}
\label{sec:story_gen}
Storytelling is at the heart of human communication, a powerful tool for connecting and conveying ideas effectively \citep{Suzuki9468}. 
It requires creativity, particularly when crafting an engaging and compelling narrative. Early approaches to this task focused on algorithmic planning based on character traits and social and physical constraints \cite{Meehan1977TALESPINAI, lebowitz1984creating}.  With the advent of powerful neural networks, the focus shifted to machine learning-based data-driven approaches \cite{du-chilton-2023-storywars, hong-etal-2023-visual-writing, akoury-etal-2020-storium, louis-sutton-2018-deep, Fan2018HierarchicalNS}. While these networks are trained on large datasets of stories and prompted to directly generate a new story, often producing locally coherent narratives, they suffer from long-term coherence, irrelevance to premise, and repetitive text problems \cite{yao2019plan}. Latest approaches have addressed these problems by using \textit{content planning} and \textit{recursive prompting} techniques where a high-level plan of the story is first generated, followed by iterative prompting that aims to generate the story in multiple steps based on the plan \cite{yao2019plan, goldfarb2020content, yang-etal-2022-re3}. Since language models are designed for open-ended text generation, controlling the attributes of its generations (e.g. topic, characters) is another major challenge \cite{Dathathri2019PlugAP}. While several methods have been developed towards \textit{controllable text generation} \cite{Dathathri2019PlugAP, Pascual2021APM, paul-frank-2021-coins, Tambwekar2018ControllableNS, Chung2022TaleBrushSS, Rashkin2020PlotMachinesOG}, language models still struggle with following \textit{constraints} \cite{sun-etal-2023-evaluating}. In addition, \textit{long-term factual inconsistency and hallucinations} still remain as major issues in language model generated texts \cite{banerjee2024llmshallucinateneedlive, Elazar2021MeasuringAI, Tam2022EvaluatingTF, Zhang2023HowLM}.

Language models have also been evaluated on their ability to produce and judge creative content as a professional writer \cite{chakrabarty2023art, Marco2024PronVP, gomez-rodriguez-williams-2023-confederacy}. \citet{chakrabarty2023art} generates short stories from LLMs based on the plots of popular fictional stories published in the New York Times and conducts a fine-grained expert assessment of both model-generated and original stories. Their study shows that LLMs significantly lag behind \textit{seasoned writers} in producing inherently creative content. Studies also demonstrate that LLMs are \textit{unreliable evaluators} of creativity \cite{Chhun2024DoLM, chakrabarty2023art}. Additionally, \cite{Tian2024AreLL} finds that LLM-generated stories are \textit{positively homogenous} and typically \textit{lack suspense} and \textit{tension}. LLMs have also been shown to produce more complex, but less creative stories than average humans \cite{ismayilzada2024evaluatingcreativeshortstory}.

To complement the shortcomings of LLMs in creative content generation, recently several works have developed frameworks to use these models as creative assistants for humans and these collaborative systems have shown strong performance across domains and editing tasks \cite{Yuan2022WordcraftSW, Schick2022PEERAC, Mirowski2022CoWritingSA, Chakrabarty2023CreativitySI, Swanson2021StoryCL}. However, recent works have also demonstrated that the output of human and language model collaboration lacks \textit{lexical} and \textit{idea diversity} \cite{Padmakumar2023DoesWW, Anderson2024HomogenizationEO}. Particularly, adapting language models with human feedback \cite{Ouyang2022TrainingLM} has been found to be a main contributing factor in diversity reduction \cite{mohammadi2024creativityleftchatprice, Padmakumar2023DoesWW, Bai2022TrainingAH}.

\subsubsection{Poetry}
\label{sec:poetry}
Poetry is a form of literary expression that uses rhythmic and often condensed creative language to evoke emotions, convey ideas, or tell stories. Early approaches to poetry generation have been based on hand-crafted templates, heuristics, and linguistic features of the target language which were limited in their expressivity \cite{Oliveira2012PoeTryMeA, Colton2012FullFACEPG, Manurung2012UsingGA, Manurung2004AnEA, manurung2000flexible, Racter1984ThePB, masterman1971, Milic1970ThePU}. However, recent statistical approaches using (recurrent) neural networks \cite{Lau2018DeepspeareAJ, ghazvininejad-etal-2016-generating, Zhang2014ChinesePG} and language models \cite{PopescuBelis2023GPoeTAL, Belouadi2022ByGPT5ES, chakrabarty-etal-2022-help, ormazabal-etal-2022-poelm, tian-peng-2022-zero, van-de-cruys-2020-automatic, agarwal2020acrosticpoemgeneration} have been shown to generate high-quality poems. While these generations almost always follow natural poetic style with appropriate rhyme and meter, they typically fail to express a \textit{poetically deep meaning} \cite{elam2023poetry, chakrabarty-etal-2023-creative}. Figure \ref{fig:poetry-ex} illustrates the qualitative difference between human and machine-generated poems. We refer the reader to \cite{Oliveira2017ASO, Elzohbi2023CreativeDG} for an in-depth survey on automatic poetry generation.

\subsubsection{Visual Creativity}
\label{sec:visual_creativity}
Humans have been producing visual content to convey emotions, concepts, and narratives since ancient times, from cave paintings and hieroglyphics to classical and Renaissance art masterpieces. For centuries, visual creativity was primarily the domain of professional artists, however, the invention of photography in the 19th century and the traditional image editing software, such as Adobe Photoshop in the past few decades enabled ordinary individuals to produce visually creative outputs without the need for formal artistic training.
The advancements of AI have further transformed the landscape of visual creativity, pushing the boundaries of what can be created and who can create it. Early works employed Generative Adversarial Networks (GAN) \cite{Goodfellow2014GenerativeAN} and Convolutional Neural Networks (CNN) \cite{lecun1989} to model images \cite{Li2016CombiningMR, Oord2016ConditionalIG, Oord2016PixelRN, Radford2015UnsupervisedRL} and \textbf{generate images} by applying specific transformations such as \textbf{style transfer} \cite{Karras2019AnalyzingAI, Abdal2019Image2StyleGANHT, Karras2018ASG, Johnson2016PerceptualLF, Dumoulin2016ALR, Gatys2015ANA}, \textbf{super-resolution }\cite{Ledig2016PhotoRealisticSI, Dong2014LearningAD}, \textbf{colorization} \cite{Zhang2016ColorfulIC} and \textbf{inpainting} \cite{Pathak2016ContextEF} or \textbf{learning a generic mapping between two images} \cite{Richardson2020EncodingIS, Huang2018MultimodalUI, Isola2016ImagetoImageTW} or \textbf{conditioning on text} \cite{Zhang2016StackGANTT, Reed2016GenerativeAT, Reed2016LearningWA, Mansimov2015GeneratingIF, Yan2015Attribute2ImageCI, Mirza2014ConditionalGA}. In recent years, the development of Transformer architecture \cite{Vaswani2017AttentionIA} and Diffusion models \cite{ho2020denoising} has further pushed AI-driven art to new heights. Trained on large amounts of multimodal data, these models are capable of generating from \textit{arbitrary} instructions not only \textbf{high-quality images} \cite{Shen2024EmpoweringVC, Geng2023InstructDiffusionAG, Saharia2022PhotorealisticTD, Ramesh2022HierarchicalTI, Gafni2022MakeASceneST, Brooks2022InstructPix2PixLT, Chakrabarty2023ISA, Patashnik2021StyleCLIPTM, Ruiz2022DreamBoothFT, Huang2022DrawYA, Hertz2022PrompttoPromptIE, Rombach2021HighResolutionIS, Nichol2021GLIDETP, Ramesh2021ZeroShotTG} but also short \textbf{photo-realistic videos} \cite{videoworldsimulators2024, Xing2023ASO, Gupta2023PhotorealisticVG, Kondratyuk2023VideoPoetAL, Luo2023VideoFusionDD, Ho2022VideoDM, Ho2022ImagenVH, Singer2022MakeAVideoTG, Arnab2021ViViTAV, Yan2021VideoGPTVG, Tulyakov2017MoCoGANDM, Vondrick2016GeneratingVW}.

Despite their impressive quality, AI systems still exhibit \textit{trivial} errors in their generations. Recent work has shown that these models struggle to effectively \textit{compose} objects with different attributes and relationships \cite{Murphy2024ACI, Huang2023T2ICompBenchAC, Zarei2024UnderstandingAM, Marcus2022AVP, Thrush2022WinogroundPV, Leivada2022DALLE2F, Conwell2022TestingRU}, fails to reliably capture common \textit{syntactic processes} such as negation, word order, comparatives etc. \cite{Leivada2022DALLE2F, Murphy2024ACI, Marcus2022AVP}, struggles with \textit{representing} numbers and texts in images \cite{Marcus2022AVP, Borji2023QualitativeFO}, often fall short when it comes to accurately depicting the intricate details of \textit{human extremities} such as hands and fingers \cite{Borji2023QualitativeFO} and lacks robust \textit{commonsense reasoning} ability \cite{Marcus2022AVP, Thrush2022WinogroundPV, Borji2023QualitativeFO, Rassin2022DALLE2IS}. Similarly, video generation models often suffer from a lack of reliable \textit{spatial reasoning}, \textit{appearance inconsistency}, \textit{temporal inalignment}, \textit{body deformation} and \textit{occlusion} issues \cite{Lei2024ACS, videoworldsimulators2024}.

\begin{figure*}[t]
\centering
\includegraphics[width=\textwidth]{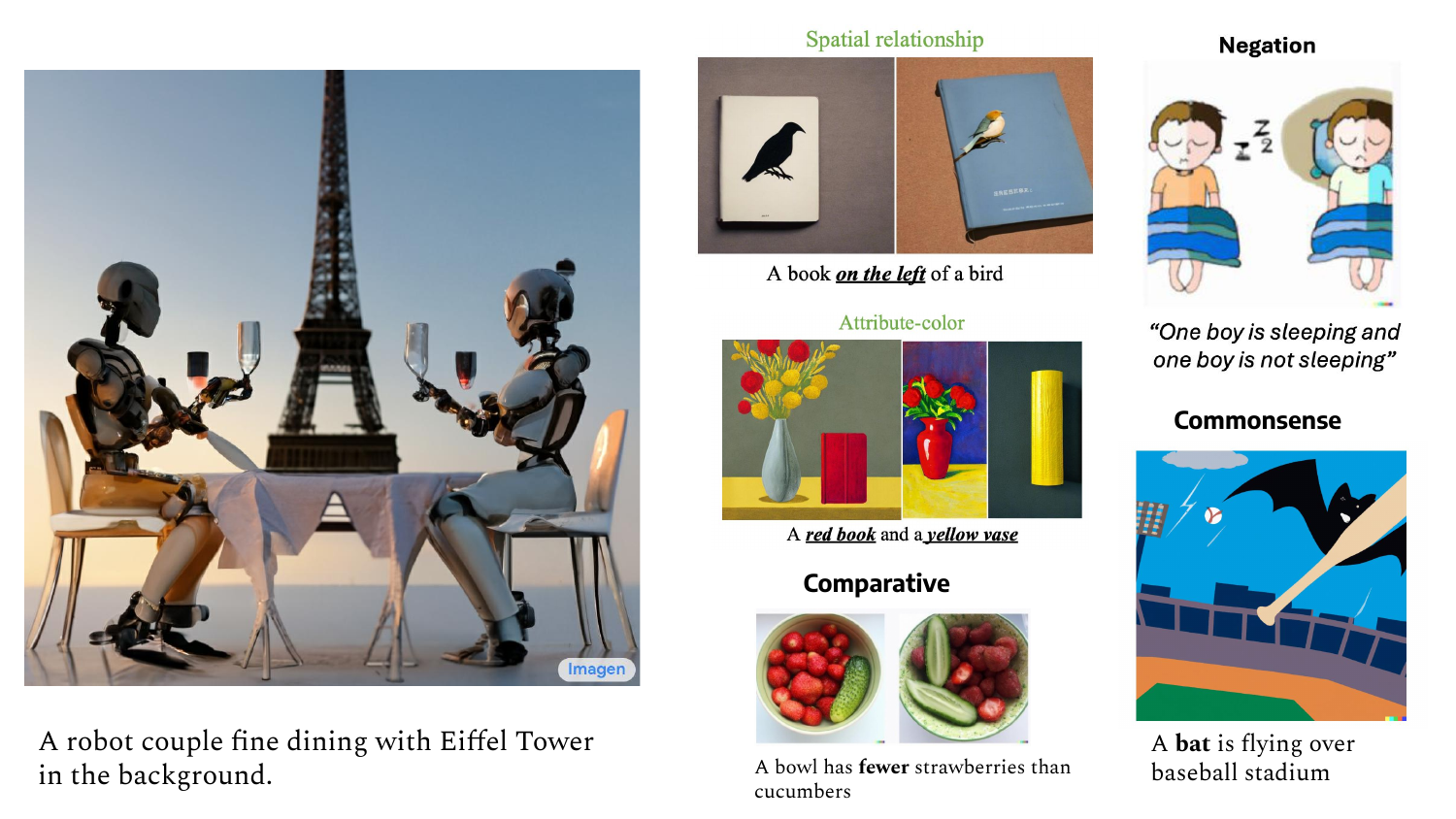}
\caption{Illustration of progress and challenges in image generation. \textbf{Left:} A creative image made by Imagen \cite{Ho2022ImagenVH} based on text instruction. \textbf{Right:} Various model generations showing failures in compositionality \cite{Huang2023T2ICompBenchAC}, commonsense \cite{Rassin2022DALLE2IS}, object relationships \cite{Marcus2022AVP}, and negation \cite{Murphy2024ACI}.}
\label{fig:visual-ex}
\end{figure*}

\subsubsection{Musical Creativity}
\label{sec:musical_creativity}
Music is another major artistic medium that allows individuals to express emotions, ideas, and cultural narratives through sound, often transcending language barriers to connect people across diverse backgrounds and experiences. Automatic music generation using computers has also a long history dating back to the 1950s \cite{shulei2024}. Early attempts at music generation employed rule-based methods \cite{hiller1953musical}, stochastic models (typically Hidden Markov Models) \cite{farbood2001analysis, ames1987automated, brooks1957experiment}, evolutionary algorithms \cite{lavrenko2003polyphonic, cope1996experiments, biles1994genjam} and recurrent neural networks \cite{eck2002finding, todd1989connectionist}. However, these methods suffered from \textit{long-range incoherence} and produced only \textit{short} pieces often with \textit{low} music quality \cite{Ji2020ACS}.

With the advent of powerful deep generative models, it became possible to capture the long-term structure of polyphonic music. Recent years have seen models that can compose multi-instrument polyphonic pieces using variational auto-encoders \cite{Roberts2018AHL, Kingma2013AutoEncodingVB}, generative adversarial networks \cite{Yang2017MidiNetAC, Dong2017MuseGANMS, goodfellow2014generative, Yu2016SeqGANSG} and transformers \cite{yuan2024chatmusician, deng2024composerx, qu2024mupt, Copet2023SimpleAC, Agostinelli2023MusicLMGM, Dhariwal2020JukeboxAG, Huang2020PopMT, Donahue2019LakhNESIM, payne2019musenet, Huang2018MusicTG}. 

While music generated by these systems often seems quite impressive, an automatic objective evaluation of music composition remains a challenge because of its subjective and complex nature \cite{Yang2018OnTE}. It is not yet entirely clear whether the AI-generated pieces are \textit{truly novel} as past work has found that deep learning-based music generation models gradually \textit{copy} increasingly distinctive chunks from pieces in the training set \cite{Yin2021AGA}. Recent studies also show that humans exhibit a preference for human compositions over AI compositions and they report something \textit{``off''} about the latter such as a lack of sense of \textit{coherence} or \textit{consistency}, \textit{odd note} choices, unnecessary \textit{complexity}, \textit{repetition, uninterestingness}, and failure to come to a \textit{resolution} \cite{Sarmento2024BetweenTA}.
\subsection{Scientific Creativity}
\label{sec:scientific-creativity}
Scientific creativity refers to the ability to generate novel ideas, approaches, or solutions within the realm of science, often leading to new discoveries, theories, or technologies. Automating the process of scientific discovery \cite{Kramer2023AutomatedSD, savage2012automating, waltz2009automate} has long been a focus of AI research dating back to the 1970s when early attempts mainly targeted \textbf{automated equation discovery} and \textbf{symbolic regression} and were based on methods such as heuristic search and genetic programming \cite{schmidt2009distilling, CoTodorovski1997DeclarativeBI, Koza1994GeneticPA, Deroski1993DiscoveringD, Rzevski1987ScientificDC, Langley1977BACONAP}. Recent methods, however, often employ Bayesian statistics \cite{guimera2020bs} and neural networks \cite{Chen2022AutomatedDO, Garcon2021DeepNN, Udrescu2020AIF2, Cranmer2020DiscoveringSM, Petersen2019DeepSR}. 

Another line of work has focused on automating the discovery of other scientific knowledge such as generating new mathematical \textbf{conjectures} or \textbf{theories} \cite{Raayoni_2021, wu2019toward, chen2016automated, BUCHBERGER2006470, FAJTLOWICZ1988113}, automatically \textbf{proving theorems} \cite{hubert2024, Trinh2024}, discovering \textbf{new concepts} \cite{HAKUK2020101080, PhysRevLett2020, lenat1984and} and predicting \textbf{new molecular structures} \cite{Abramson2024, zambaldi2024novodesignhighaffinityprotein, Jumper2021, lindsay1980applications} among others. A notable example in this area is the recent AlphaFold model \cite{Jumper2021} that can predict millions of intricate 3D protein structures which has the potential to significantly accelerate research in biology.

While the above works have mainly targeted one aspect of the \textit{scientific process}, namely the automatic discovery of particular scientific knowledge, there have also been attempts to partially or fully automate the \textit{entire} process itself. The scientific process typically starts with a \textit{scientific question} or an \textit{idea} that is then used to formulate a \textit{hypothesis}, followed up with  \textit{designing and running experiments} and \textit{analyzing results} to test the validity of the hypothesis and ends with \textit{communicating} the findings to the scientific community \cite{Kramer2023AutomatedSD}. Recently, the field has seen a surge in the development of frameworks using neural networks and especially, large language models to automate several steps of the scientific process such as \textbf{literature review} \cite{skarlinski2024language}, \textbf{idea generation} \cite{si2024llmsgeneratenovelresearch, Baek2024ResearchAgentIR, wang2024scimonscientificinspirationmachines, castelo2024ai, Girotra2023IdeasAD, Wang2023SciMONSI},  \textbf{hypothesis generation} \cite{wang2024scimonscientificinspirationmachines, Ghafarollahi2024SciAgentsAS, majumder2024discoverybenchdatadrivendiscoverylarge, qi2023largelanguagemodelszero, Yang2023LargeLM, Sybrandt2020} and \textbf{paper writing} \cite{altmae2023, Wang2019PaperRobotID}. Yet other works have gone further to introduce a so-called \textbf{``AI Scientist''} that automates almost the entire scientific process from the idea generation to experiment execution to even paper writing \cite{lu2024aiscientistfullyautomated, ifargan2024autonomousllmdrivenresearchdata, Li2024MLRCopilotAM, liu2024coquest, boiko2023emergentautonomousscientificresearch, king2009automate}.

While the latest advancements in the automation of scientific creativity are remarkable, these results should be taken with a grain of salt. Most of the recent end-to-end automation frameworks are powered by LLMs, hence, they face the same challenges and issues we discussed in the previous sections such as hallucinations, lack of content diversity, novelty, and robust reasoning capabilities. For example, one of the aforementioned large-scale idea generation studies \cite{si2024llmsgeneratenovelresearch} finds that out of $4,000$ LLM-generated ideas only $200$ are unique. Their qualitative analysis also reveals some common failure modes such as \textit{vague} implementation details, \textit{misuse} of datasets, \textit{inappropriate} baselines, \textit{unrealistic} assumptions, and overall \textit{poorly-motivated} ideas. Similarly, another study benchmarking the machine learning experimentation capabilities of LLMs reports \textit{hallucinations} and \textit{poor planning} as some of the major issues with these models \cite{huang2024mlagentbenchevaluatinglanguageagents}.

Another important aspect of scientific discovery is the \textbf{explainability} \cite{Li2021FromKT} which helps humans prevent or better prepare for a possible future technological singularity \cite{Good1965SpeculationsCT, Ulam1958JohnVN}. However, current LLMs are largely black-box AI systems, and allowing them to make discoveries that are incomprehensible to humans may lead to a scenario where human knowledge is left far behind the machine's knowledge resulting in machines that humans can't control \cite{Good1965SpeculationsCT}.

\section{Creativity and Copyright}
\label{sec:copyright}
Our brief survey into the methods used to produce creative outputs showed that the predominant approach is currently the generative deep learning techniques, especially LLMs. These models typically have billions of adjustable parameters \cite{brown2020language} and are trained on massive amounts of public and private data \cite{raffel2023exploringlimitstransferlearning}. Consequently, these models have been found to exhibit strong memorization skills \cite{Carlini2022QuantifyingMA, Carlini2018TheSS} such that they can sometimes copy large passages \cite{chang2023speakmemoryarchaeologybooks, McCoy2021HowMD} or replicate images from their training data \cite{Somepalli2022DiffusionAO}. While this could be of little concern when the duplicated content is public and generic, however, the training datasets of popular LLMs are often undisclosed and can include private and copyrighted data leading to concerns about \textbf{copyright infringement and privacy violation} \cite{Franceschelli2021CopyrightIG}. Although several approaches have been developed to \textit{detect} \cite{duarte2024decopdetectingcopyrightedcontent, shi2024detectingpretrainingdatalarge, li2024diggerdetectingcopyrightcontent, oren2023provingtestsetcontamination, carlini2021extractingtrainingdatalarge} and \textit{prevent} \cite{Hans2024BeLA, zhao2022provablyconfidentiallanguagemodelling, Kandpal2022DeduplicatingTD, Ippolito2022PreventingVM} unintended memorization in LLMs, major questions concerning the use of copyrighted material for training and \textit{authorship} of the machine-generated content remain unresolved \cite{abbot2023, Franceschelli2021CopyrightIG}. Recent lawsuit between The New York Times and OpenAI \cite{lawsuitNYTOpenai} and the class action\footnote{A class action is a type of civil lawsuit brought on behalf of many similarly situated people who have been harmed in the same way by the same entity.} against Stable Diffusion, Midjourney, and DeviantArt \cite{classactionstability} have further highlighted the urgency of the matter and the need for clear legal frameworks that address the complex issues surrounding intellectual property rights, ethical use, and the boundaries of fair use in AI development. 

More specifically, two key questions concerning copyright and authorship are of interest and here we briefly discuss them with respect to machine-generated artworks. We refer the reader to \citet{Franceschelli2021CopyrightIG} and \citet{abbot2023} for a detailed discussion of these questions.

\paragraph{\textbf{Is it copyright infringement to use protected works for the training of generative models?}}

To answer this question, we will review the implications of the existing relevant laws from the US and EU. Under the US Law Code, reproduction of a copyrighted work can be allowed if the use can be considered a \textit{fair use} of the work \cite{Netanel2011MakingSO}. Analyzing the criteria used to determine fair use, \citet{Franceschelli2021CopyrightIG} concludes that it is not straightforward to assess this for generative deep models and if these models do not add any form of novelty to their output. Their outputs may not qualify for fair use, which can potentially derail the progress in AI \cite{Sobel2017ArtificialIF}.

Under the EU law, on the other hand, the use of lawfully accessible protected work for training is permitted as long as 1) the rightsholder of the used data has not reserved the right to withhold its data from being reproduced and 2) the accessed data is retained only for the time required for the purposes of scientific research \cite{Franceschelli2021CopyrightIG}. However, \citet{Franceschelli2021CopyrightIG} also notes while the second criterion is reasonably easy to satisfy, the first criterion is hard to verify in practice because nowadays, models are being trained on large amounts of data published on the internet for which there is no centralized repository allowing to filter \textit{reservation-free} works. Finally, whether providers of such a repository or the developers of the models should be forced to perform this check is unclear.

\paragraph{\textbf{Who is the author (if someone) or who will own the copyright on the generated work?}}
To answer this question, first, we have to make a distinction between the \textit{AI-assisted} and \textit{AI-generated} content. If the generative model is merely used as a tool to \textit{assist} a human to produce a creative artwork, then the human will be considered the author and own the copyright. However, it becomes tricky to determine the authorship and the copyright status of the work that is \textit{generated} mostly by AI with little human involvement (e.g. human as prompter). First, let's consider the authorship issue. Some have argued that for an author to exist there has to be a message that the author wants to convey through their work, but since no one can reliably predict the output of a generative model, \textit{no author} exists \cite{Ginsburg2018PeopleNM}. However, if we suppose that there is an author, then there are mainly three contenders in question: 1) the person who developed the AI model (\textit{developers}) 2) the person who used the AI model to produce creative work (\textit{users}), and finally 3) the \textit{AI model} itself. Since the existing laws in most countries only attribute copyright to a human, but not to a machine, the main tension is around deciding whether to attribute the authorship (also the copyright) to users or developers \cite{abbot2023, Santos2020IntellectualPO, Deltorn2018AuthorshipIT, Guadamuz2017DoAD}. Some have argued that the criterion to determine authorship should center around the incentives to create and promote the work, not the ideation and creation of the work itself \cite{Miller1993COPYRIGHTPF} and since the \textit{users} of the generative models are best positioned to do so, they should be assigned the authorship \cite{Samuelson1986AllocatingOR, Denicola2016ExMC, Franceschelli2021CopyrightIG}. Another argument supporting this assignment is by ruling out the developer as the author since they just create the \textit{potentiality} for the creation of the output, but not its \textit{actuality} \cite{Franceschelli2021CopyrightIG, Samuelson1986AllocatingOR}. Using the analogy proposed by \citet{ralston2005}, it would be similar to claim a knife manufacturer is more responsible for murder than the person who wielded the knife or assigning copyright to the teacher of the painter rather than the painter himself/herself \cite{Franceschelli2021CopyrightIG}. Finally, arguments in favor of AI authorship have also been made recently suggesting that this will promote transparency, efficient allocations of rights, and even counterintuitively protect human authors \cite{abbot2023}.
\section{Future Directions}
\label{sec:future_directions}
In the previous sections, our brief exploration into the creativity of modern AI systems revealed that these systems exhibit some capacity for producing linguistically and artistically creative outputs and thinking creatively. However, true human-like creative abilities seem to be still out of reach, as indicated by challenges with tasks demanding creative problem-solving \cite{Jiang2023BRAINTEASERLT, Tian2023MacGyverAL}, abstract reasoning \cite{Mitchell2023ComparingHG, Gendron2023LargeLM}, and compositionality \cite{Huang2023T2ICompBenchAC, Murphy2024ACI}. Some studies also highlighted major issues in machine outputs, such as lack of originality \cite{lu2024aihumanityssalieriquantifying, chakrabarty2023art}, diversity \cite{Anderson2024HomogenizationEO, Padmakumar2023DoesWW} and incoherence \cite{Tam2022EvaluatingTF, Sarmento2024BetweenTA}. From the Four-C model perspective, these models seem to manifest only mini-c or little-c type of creativity while Pro-C and Big-C creativity remain elusive. Similarly, current AI models exhibit strong interpolation and moderate extrapolation capabilities. However, they are still far from truly \textit{inventing} a completely new type of creative artefact. In this section, we discuss potential research directions that can help us better measure and improve the creative abilities of AI systems.

\subsection{Evaluating Creativity}
\subsubsection{Creative Process}
Cognitive scientists and psychologists have proposed theoretical frameworks to evaluate creativity such as characterizing it based on \textbf{\textit{input}}, \textbf{\textit{process}} and \textbf{\textit{output}} \cite{Jordanous2012ASP, pease2002evaluate, Ritchie2007SomeEC} or four Ps: \textbf{\textit{person}}, \textbf{\textit{product}}, \textbf{\textit{process}} and \textbf{\textit{press}} \cite{Jordanous2016TheLT, rhodes1961fourp}. A common thread across all these theories is their emphasis on evaluating the \textit{process} aspect of creativity. 
However, most works in AI, including the ones we reviewed before, focus on evaluating and analyzing creativity from the \textit{output} or \textit{product} perspective. Creative \textit{process}, on the other hand, is an equally (or perhaps more) important aspect of creativity that can tell us how creativity \textit{``arises''} in the first place and what the key ingredients involved \cite{Colton2008CreativityVT}. For example, in computational creativity, one popular theory by 
\citet{boden2004creative} defines the creative process in terms of manipulations over a conceptual space. This theory divides creativity into three types: \textbf{combinatorial} that makes unfamiliar connections between familiar concepts (e.g. creating hybrid
fictional creatures such as pegasus, sphinx, or mermaid), \textbf{exploratory} that involves an open-ended search in a conceptual space (e.g. a novel chess move) and \textbf{transformational} that requires a fundamental transformation of the existing conceptual space (e.g. non-Euclidean geometry\footnote{https://en.wikipedia.org/wiki/Non-Euclidean\_geometry}). Another popular theory by \citet{wallas1926art} explains the creative process in four stages akin to how scientists develop their ideas: \textbf{preparation} stage where the problem at hand is investigated in all directions, information is gathered and analyzed, \textbf{incubation} stage where you step back from the problem and let your unconscious work through it in the background, \textbf{inspiration} stage where a creative insight is typically realized (an ``Aha!'' or ``Eureka!'' moment) and finally, \textbf{verification} stage where you test, evaluate and build further on your creative idea to make it perhaps useful.

While AI systems produce seemingly creative outputs, the nature of the creative process they employ (if any) remains unknown. Only very recently attempts have been made to study the creative process of machines \cite{nath2024characterisingcreativeprocesshumans} which analyzes the creative process of language models and humans to solve AUT task using \textbf{\textit{response pathways}} (persistent vs. flexible) \cite{baas2013, Nijstad2010TheDP} and finds that while humans are able to follow a mixture of pathways, models are biased towards either one of them pointing to a limited capacity. Hence, analyzing the creative process of machines is an emerging and exciting area for which much work remains to be explored. We believe a strong collaboration between the computational creativity \cite{veale2019computational, Colton2012ComputationalCT} and NLP communities drawing ideas from past research on studying human creative process and techniques from research on \textit{(mechanistic) interpretability}\footnote{https://www.neelnanda.io/mechanistic-interpretability} \cite{saphra2024mechanistic, Bereska2024MechanisticIF} could lead to a better understanding of the creative capacity of AI systems.

\subsubsection{Dimensions of Creativity}
As we discussed earlier, there are many dimensions of creativity, but most works generally focus on evaluation of the \textit{novelty} and \textit{usefulness} dimensions. However, \textit{surprise}, \textit{agency} and \textit{spontaneity} dimensions are also equally important. Humans typically communicate an emotion or a deeper meaning through creative products and their creative process is characterized by spontaneous ``Aha'' or ``Eureka'' moments coupled with deliberate decisions made at each step of the way \footnote{https://www.newyorker.com/culture/the-weekend-essay/why-ai-isnt-going-to-make-art}. However, current AI systems lack agency and are typically trained to generate the most likely output leaving no room for any intentional or spontaneous action \cite{peeperkorn2023characterizations, franceschelli2023creativitylargelanguagemodels}. Therefore, a holistic evaluation of machine creativity should involve consideration of all these different dimensions that characterize human creativity.

\subsection{Improving Creativity}
Recent years have seen a surge in human-AI creative collaboration \cite{Vinchon2023ArtificialI} popularized by the introduction of chat-based products such as ChatGPT\footnote{https://chat.openai.com} and Gemini\footnote{https://gemini.google.com}. However, the poor creative capacity of current AI systems necessitates the innovation of new techniques to improve the creativity of their outputs. In this section, we discuss several possible directions to take.

\subsubsection{Creative Architectures}
As we argued before, current AI architectures optimized for the most likely outcome might have fundamental limitations to exhibit true human-like creativity. In fact, by definition, current AI models are optimized to model the training distribution while creating something new requires the model to \textit{diverge} from its learned distribution. Therefore, innovating at the architecture level to endow machines with mechanisms to actively diverge from the training data and a capacity for \textit{agency} and \textit{spontaneity} might be a necessary step towards robust creativity. An emerging new research area called \textbf{\textit{active divergence}} attempts to optimize models for creativity using methods such as novelty search, divergent fine-tuning, and objective functions targeting different dimensions of creativity \cite{Broad2021ActiveDW, elgammal2017can, Guimaraes2017ObjectiveReinforcedGA, Bunescu2019LearningTS}.

\subsubsection{Creative Prompt Engineering}
Natural language-based interaction with the current AI systems has created an intuitive playground to elicit more capabilities from these systems \cite{Qiao2022ReasoningWL}. These so-called \textit{prompt engineering} techniques have also been shown to enhance the creativity of large language models \cite{Nair2024CreativePS, Mehrotra2024EnhancingCI, Tian2023MacGyverAL, SummersStay2023BrainstormTS}. We can draw ideas from psychology that has shown techniques such as \textit{brainstorming} \cite{Osborn1957AppliedI}, \textit{competence injection} \cite{Liu2020FeelingOC} and \textit{threatening situations} \cite{riley2019evidencethreateningsituationsenhance} stimulate creativity of humans. Hence, designing prompts inspired by these methods is a promising direction to get the most out of future AI systems.

\subsubsection{Creative Decoding}
An important component in natural language generation is the decoding strategy which is a significant contributor to the quality of the generation \cite{meister-etal-2022-high}. Past work has shown that simple greedy decoding results in repetitive and uninteresting generations \cite{li2023contrastive} and numerous powerful decoding algorithms have been developed to address these problems \cite{Holtzman2019TheCC, Fan2018HierarchicalNS, meister-etal-2023-locally}. These decoding strategies mainly target generating human-like text and do not directly target creativity. A popular approach is to increase the randomness of the output by increasing the \textit{temperature} parameter, however, recent work shows that this parameter is weakly correlated with the novelty of the output \cite{peeperkorn2024temperaturecreativityparameterlarge}. A potential direction could be to devise new creative decoding algorithms that go beyond the temperature parameter by injecting \textbf{\textit{semantic planning}} or intentionality \cite{Franceschelli2024CreativeBS} and employing information-theoretic measures of novelty, utility, and surprise \cite{bunescu-uduehi-2022-distribution, kuznetsova-etal-2013-understanding, heinen2017semantic}.
\section{Conclusion}
\label{sec:conclusion}
In conclusion, while the rapid advancements in AI, particularly through state-of-the-art models such as large language models, diffusion models, etc., have demonstrated impressive capabilities in generating creative outputs, the question of genuine machine creativity remains unresolved. This survey has explored key areas of linguistic creativity, creative problem-solving, and artistic and scientific creativity, providing a comprehensive overview of the state of AI creativity. We also discussed pressing copyright and authorship issues with generative artworks, highlighted major challenges facing current AI systems and proposed potential research directions on how to evaluate and improve the creativity of these systems. We believe our suggestions can help future research to determine if machines can achieve a human-like creative process, ultimately enriching our understanding of artificial intelligence and its capabilities.
\begin{acks}
We gratefully acknowledge the support of Swiss National Science Foundation (grant 205121\_207437: C - LING). We also thank Angelika Romanou for her help with creating figures for this paper.
\end{acks}

\bibliographystyle{ACM-Reference-Format}
\bibliography{literature}

\end{document}